\documentclass{article}
\usepackage{iclr2027_conference,times}

\usepackage{amsmath,amsfonts,bm}

\def\eqref#1{equation~\ref{#1}}
\def\1{\bm{1}}

\DeclareMathAlphabet{\mathsfit}{\encodingdefault}{\sfdefault}{m}{sl}
\SetMathAlphabet{\mathsfit}{bold}{\encodingdefault}{\sfdefault}{bx}{n}

\usepackage{xcolor}
\usepackage{hyperref}
\usepackage{url}
\usepackage{xurl}
\usepackage{graphicx}
\usepackage{booktabs}
\usepackage{multirow}
\usepackage{makecell}
\usepackage{enumitem}
\usepackage{array}
\usepackage{float}
\usepackage{needspace}

\newcommand{\benchname}{FabricBench}

\newlength{\figfourimageheight}
\newlength{\figfourpanelheight}

\newcommand{\cmark}{$\checkmark$}
\newcommand{\xmark}{$\times$}

\newcommand{\scr}[1]{\,{\scriptsize\textcolor{gray}{(#1)}}}
\newif\ifrevhighlight
\revhighlightfalse
\ifrevhighlight
  \hypersetup{pdfborder={0 0 0}}
  \newcommand{\rev}[1]{{\color{blue}#1}}
  \newcommand{\revon}{\color{blue}}
\else
  \newcommand{\rev}[1]{#1}
  \newcommand{\revon}{}
\fi

\title{FabricBench: A Standardized Policy Benchmark for Bimanual Fabric Manipulation with a Contact-Driven Deformable Simulator}

\author{%
  \textbf{Wenkang Hu\textsuperscript{1,*}\quad Haoran Wang\textsuperscript{1,*}\quad Liu Liu\textsuperscript{2}\quad Wenchao Huang\textsuperscript{3}\quad Yiji Chen\textsuperscript{1}} \\[1pt]
  \textbf{Mengao Zhao\textsuperscript{2}\quad Xincheng Tang\textsuperscript{1}\quad Zhengjie Shu\textsuperscript{1}\quad Lai Jiang\textsuperscript{1}\quad Wei Li\textsuperscript{1}} \\[1pt]
  \textbf{Zhendong Wang\textsuperscript{3}\quad Zhizhong Su\textsuperscript{2}\quad Huamin Wang\textsuperscript{3}\quad Ruigang Yang\textsuperscript{1,\dag}} \\[4pt]
  \normalfont\small \textsuperscript{1}Shanghai Jiao Tong University\quad
  \textsuperscript{2}Horizon Robotics\quad \textsuperscript{3}Style3D Research \\[2pt]
  \normalfont\footnotesize \textsuperscript{*}These authors contributed equally.\qquad
  \textsuperscript{\dag}Corresponding author.}
\iclrfinalcopy
\begin{document}

\maketitle
\pagestyle{plain}

\begin{abstract}
Fabric manipulation challenges robot policies with changing geometry,
self-occlusion, and contact, yet standardized evaluation remains focused
on rigid objects. We introduce \benchname{}, a benchmark for bimanual
fabric manipulation with six tasks spanning garments, silk, and bags,
10{,}000 simulation-ready assets, and controlled tests of configuration
and appearance generalization. Built on an open-source solver in Newton, \rev{our simulator enables stable bimanual grasping, holding, and release through contact forces, without artificial attachment. Redesigned cloth--rigid contact and bidirectional force coupling support interactions among robot arms, fabric, and free rigid objects, allowing a bag to support its contents and transmit their weight to the grippers. In a controlled grasp--hold--release evaluation, our solver succeeds in all 45 trials, compared with none for the tested baseline configurations.} Evaluating six representative policies,
including vision-language-action models and world action models, reveals
substantial remaining challenges: the best policy achieves 77\% average
success, while no policy exceeds 36\% on long-horizon T-shirt folding.
Under the most deformed initial T-shirt configurations, success falls
to 6--14\% across policies, with larger drops than under the evaluated
appearance shifts. Zero-shot deployment of $\pi_{0.5}$ on a real robot
largely preserves task difficulty ordering across six tasks
(Spearman $\rho=0.94$). These results establish \benchname{} as a
reproducible testbed for measuring fabric-manipulation capabilities
and exposing limitations in configuration generalization.
The project website is available at \url{https://fabricbench.github.io/}.
\end{abstract}

\section{Introduction}
\begin{figure}[t]
\centering
\includegraphics[width=\textwidth]{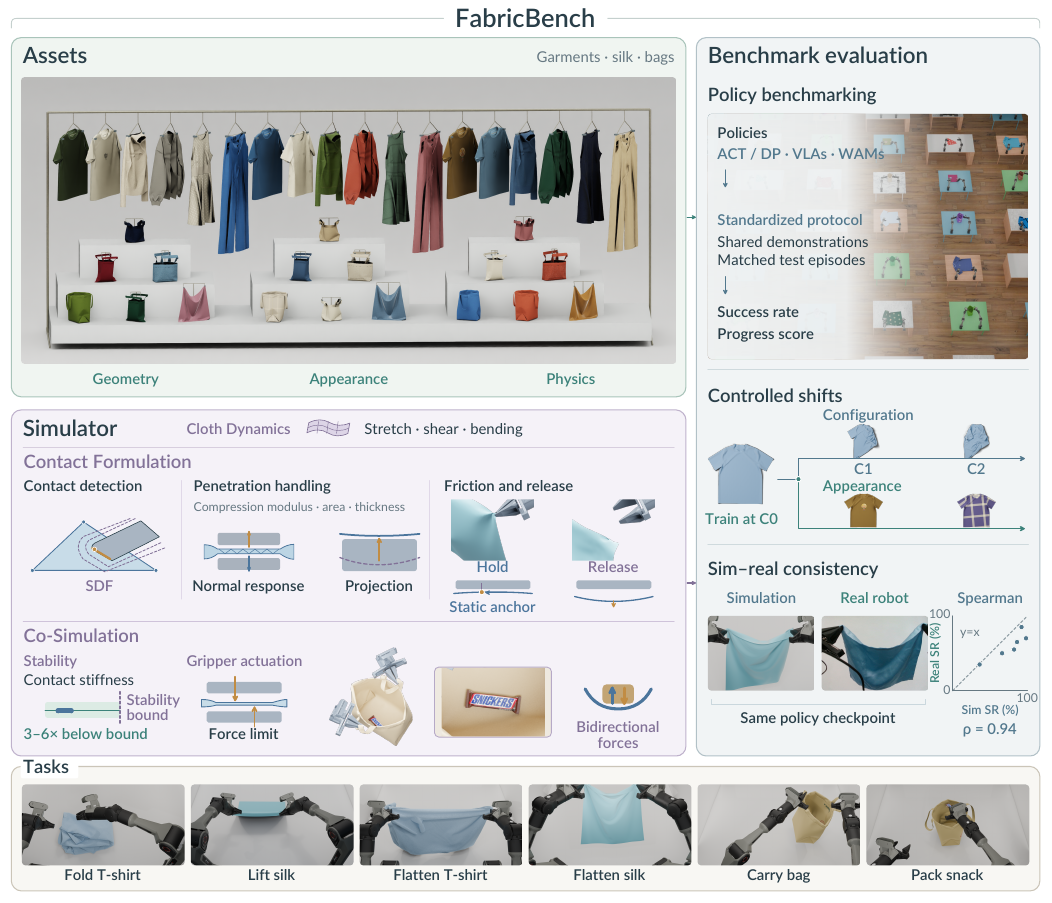}
\caption{\textbf{Overview of \benchname{}.}
\textbf{Assets (top left):} garments, silk, and bags with geometry, appearance, and physical properties.
\textbf{Simulator (bottom left):} contact detection, penetration handling, frictional grasp and release,
force-limited grippers, and bidirectional coupling make grasping, release, and load transfer
arise from contact forces without artificial attachment.
\textbf{Benchmark evaluation (right):} shared demonstrations and matched test episodes;
separate configuration and appearance tests, with geometry and physics fixed for appearance shifts;
sim--real task difficulty comparison using the same checkpoint per task.
\textbf{Tasks (bottom):} six tasks spanning long-horizon folding, configuration recovery,
friction-limited grasping, and load-bearing \mbox{fabric--object interactions}.}
\label{fig:teaser}
\end{figure}

\label{sec:intro}

Fabric manipulation is central to everyday tasks such as folding laundry and carrying bags, yet folds, self-occlusion, and changing contacts make it difficult to evaluate reliably~\citep{lin2020softgym,lu2024garmentlab}. Vision-language-action (VLA) models and world action models (WAMs) show progress on fabric tasks, including laundry folding with $\pi_0$~\citep{black2024pi0} and towel folding with Fast-WAM~\citep{yuan2026fastwam}, while standardized manipulation benchmarks remain focused on rigid objects~\citep{mu2025robotwin,liu2023libero}. We introduce \benchname{}, a standardized benchmark for evaluating learned policies on bimanual fabric manipulation.

Simulation can make fabric evaluation scalable and reproducible, but reliable cloth contact remains challenging. Existing environments report penetration, jitter, and instability~\citep{wang2025dexgarmentlab,flatlab2026}; others use attachment rules to bind cloth to the gripper~\citep{lu2024garmentlab,flash2026}. Under learned policies, small fluctuations in continuous gripper commands can repeatedly trigger attachment and release, making outcomes depend on simulator rules rather than policy capability.

To address this limitation, we develop an improved open-source cloth solver based on the Style3D cloth solver in the Newton physics engine. Redesigned cloth--rigid contact and bidirectional force coupling among the arm, fabric, and free rigid objects allow grasping, release, and load transfer to arise from contact and remain stable under learned policies. Six tasks cover T-shirt folding and flattening, silk flattening and lifting, bag carrying, and packing, spanning multilayer self-contact, dynamic manipulation, self-occlusion, and load-bearing fabric--object interactions.

Beyond task success, \benchname{} separates generalization across initial fabric configurations from generalization across appearance. Folding or crumpling the same garment changes visibility, contacts, and feasible actions. For the two flattening tasks, seeded configuration levels (C0--C2) vary initial deformation while sharing assets and placement distributions. Held-out appearance variants instead change color and texture while keeping geometry and physics fixed. These controlled axes test whether policies cope with changes in fabric state as well as visual appearance.

This paper makes the following contributions:
\begin{itemize}[leftmargin=1.2em,itemsep=1pt,parsep=0pt,topsep=2pt]
\item \textbf{A fabric manipulation benchmark.} \benchname{}: six tasks over garments, silk
cloth, and fabric bags; held-out variations in initial deformation and appearance on the
flattening tasks; a
unified training and evaluation protocol with demonstration generation.
\item \textbf{An open-source simulator.} An improved cloth solver based on the Style3D solver in the
Newton physics engine, with redesigned cloth--rigid contact and bidirectional force coupling among the arm, the fabric, and free
rigid objects, so that grasping,
release, and load transfer arise from contact forces and remain stable under learned policies.
\item \textbf{A systematic assessment of policy limitations.}
Our evaluation shows that current policies remain unreliable on fabric
manipulation, especially long-horizon folding. Appearance shifts reduce
performance, although some policies show partial robustness to these
changes. On the flattening tasks, performance degrades systematically
with increasing initial deformation, with drops far exceeding those
under appearance shift.
\end{itemize}

\section{Related Work}

\subsection{Manipulation Policies and Benchmarks}
\label{sec:rw-bench}

Manipulation policies have progressed rapidly from classical imitation
learning~\citep{zhao2023act,chi2023diffusionpolicy} to large-scale generalist models.
Vision-language-action models such as OpenVLA~\citep{kim2024openvla}, $\pi_0$,
$\pi_{0.5}$~\citep{black2024pi0,pi052025}, and GR00T~\citep{bjorck2025gr00t} are pretrained on
large multi-robot datasets and improve instruction following and cross-embodiment transfer, and
world action models such as LingBot-VA~\citep{li2026lingbotva} and
Fast-WAM~\citep{yuan2026fastwam} further predict future observations jointly with actions.

Standardized simulation benchmarks such as RoboTwin~\citep{mu2025robotwin,chen2025robotwin2}, LIBERO~\citep{liu2023libero}, RoboDojo~\citep{robodojo2026}, and RoboLab~\citep{yang2026robolab} have driven this progress on rigid bodies through fixed task suites, public demonstrations, and shared protocols. Studies of policy degradation under object-layout, camera-viewpoint, and initial-pose perturbations~\citep{liberopro2025,liberoplus2025} have further exposed where these policies break and guided their improvement. Deformable objects are absent from these suites or appear as a single task~\citep{robodojo2026}, and the deformation state of the object lies outside their perturbation axes.

Early benchmarks for deformable objects, SoftGym~\citep{lin2020softgym},
DaXBench~\citep{chen2023daxbench}, and Deformable Ravens~\citep{seita2021deformable}, introduced
simulated task suites for cloth, rope, and bags, but observe through low-resolution renders,
heightmaps, or particle states, whereas simulated scores track real-robot performance once the
visual gap is controlled~\citep{li2024simplerenv}. Recent Isaac Sim suites such as
LeHome~\citep{lehome2026} and FlatLab~\citep{flatlab2026} narrow this visual gap for specific
categories of deformable objects. \benchname{} further narrows the physics gap, bringing simulated
evaluation of deformable manipulation closer to the real world.
\rev{A detailed comparison is given in Appendix~\ref{app:bench-compare}.}

\subsection{Cloth Simulation for Robotic Manipulation}
\label{sec:rw-sim}

Position-based dynamics (PBD/XPBD)~\citep{muller2007pbd,macklin2016xpbd} underlies many
cloth-manipulation environments: SoftGym builds on the PBD-based FleX
engine~\citep{macklin2014unified}, and GarmentLab~\citep{lu2024garmentlab},
DexGarmentLab~\citep{wang2025dexgarmentlab}, LeHome, and FlatLab use the PBD cloth of Isaac Sim~\citep{isaacsim}. PBD is stable and
easy to implement, but it approximates continuum mechanics with geometric constraint
projections, producing over-stretching and self-penetration, and its gap to real garment
dynamics has been measured~\citep{blancomulero2024simtoreal,hu2026rgbench}. Finite-element and
intersection-free solvers~\citep{faure2012sofa,li2020ipc,li2021cipc}, including the cloth solver
of Genesis~\citep{genesis2024}, are more accurate but about an order of magnitude slower,
too slow for policy-scale rollouts (Appendix~\ref{app:sim-criteria}), whereas recent fast FEM cloth solvers reach continuum-level
accuracy at practical rates~\citep{hu2026rgbench,flash2026}. A simplified open-source version of
the Style3D solver following this principle is available in
the Newton physics engine~\citep{newton2025,nvidia2025newtonblog}, and we build on it.

Contact between cloth and rigid bodies, such as a gripper or objects in a bag, is typically handled
by rules or artificial forces rather than simulated physically: SoftGym grasps with pickers that
attach nearby particles, DaXBench with pick-and-place macro actions, and Deformable Ravens with
fixed constraints; GarmentLab~\citep{lu2024garmentlab} and FLASH~\citep{flash2026} bind cloth vertices within a
distance threshold when the gripper closes, and DexGarmentLab~\citep{wang2025dexgarmentlab} adds artificial
adhesion between the fingers and the fabric; the released Style3D solver~\citep{newton2025}
couples cloth and rigid bodies only one way, so objects in a bag follow it only if
a rule binds them to it. In all these cases, whether the cloth or the bag's contents follow the
gripper is decided by these mechanisms rather than by contact forces, and, as discussed in
Section~\ref{sec:intro}, end-to-end policies have no access to them. In \benchname{}, grasp,
release, and load transfer arise from bidirectional contact forces alone, so task outcomes reflect the
policy's actions under real physical contact.

\section{Deformable Simulation and Contact Formulation}
\label{sec:platform}

We state the cloth dynamics, the contact energy between cloth and rigid bodies, and
the force exchange with the rigid-body solver; implementation details are in
Appendix~\ref{app:sim}.

\subsection{Cloth Dynamics}
\label{sec:cloth}

\paragraph{Material model.} A garment is a triangle mesh. Its in-plane energy is the
anisotropic membrane model of \citet{baraff1998large}, with warp, weft, and shear
stiffness specified per triangle; its out-of-plane energy is the discrete bending
model of \citet{bergou2006quadratic} on mesh edges. These stiffnesses and the areal
density correspond directly to standard fabric tests, so they are loaded from
measured values and carry no non-physical tuning parameters.

\paragraph{Time integration.} Positions advance by implicit Euler with a fixed substep
$\Delta t$. The positions at the end of a substep minimise an incremental
potential,
\begin{equation}
\mathbf{x}_{n+1} = \arg\min_{\mathbf{x}}\;
\frac{1}{2\Delta t^{2}}\,\|\mathbf{x}-\tilde{\mathbf{x}}\|_{\mathbf{M}}^{2}
+ E_{\mathrm{el}}(\mathbf{x}) + E_{\mathrm{c}}(\mathbf{x}),
\label{eq:potential}
\end{equation}
where $\tilde{\mathbf{x}}$ is the inertial prediction, $E_{\mathrm{el}}$ the
membrane and bending energy above, and $E_{\mathrm{c}}$ the contact energy of
Sec.~\ref{sec:contact}. The minimisation uses projective-dynamics
iterations~\citep{bouaziz2014projective}: the Hessian of $E_{\mathrm{el}}$ is
replaced by a constant approximation assembled once, the gradient and diagonal
Hessian of $E_{\mathrm{c}}$ are re-evaluated at every iteration, and each linear
system is solved by Jacobi-preconditioned conjugate gradients; self-contact
repulsion and untangling terms~\citep{volino2006} are part of $E_{\mathrm{c}}$. The
whole substep runs as one replayed CUDA graph with fused conjugate-gradient kernels
and table-based distance queries (Appendix~\ref{app:sim-gpu}).

\subsection{Contact Formulation}
\label{sec:contact}

For fixed friction anchors, the contact potential between the cloth and the gripper fingers is
\begin{equation}
E_{\mathrm{c}}(\mathbf{x}) = \sum_{c}\frac{k_c}{2}\big(h - d_c(\mathbf{x})\big)_{+}^{2}
+ \sum_{c}\frac{k_t}{2}\big\|\mathbf{P}_c\big(\mathbf{x}_c - \mathbf{p}_c\big)\big\|^{2},
\qquad d_c(\mathbf{x}) = \min_{\mathbf{y}\in\triangle_c}\phi(\mathbf{y}),
\label{eq:contact}
\end{equation}
over active contacts $c$, combining a normal penalty and an elastic tangential trial
potential. This compliant-shell formulation uses no barrier term or continuous
collision detection. Relative to released cloth stacks it changes three things.

\paragraph{Contact detection.} The cloth is a shell of half-thickness $h$
(Appendix~\ref{app:sim-contact}). In Eq.~(\ref{eq:contact}), $d_c$ is
the minimum over the \emph{whole triangle} of the rigid body's precomputed signed
distance field $\phi$, not a vertex distance; a contact is active when $d_c < h$.
The minimisation is seeded at the triangle's vertices and centroid and at the
intersections of the rigid mesh's edges with the triangle, so that a sharp finger
edge passing through the interior of a triangle is detected. With the triangle as
the contact primitive, $h$ can be chosen below the mesh edge length, whereas
vertex-based contact must set its collision radius above it to stop a finger from
passing between vertices, tying the effective thickness to mesh resolution. The forces of Eq.~(\ref{eq:contact}) also act on the
rigid body (Sec.~\ref{sec:cosim}). The table and free objects, which are not
pinched, keep the solver's standard vertex-based penalty contact.

\paragraph{Penetration handling.} The normal stiffness is derived from material and
geometry, $k_c = E_t A_c / t_0$, with $E_t$ the fabric's transverse compression
modulus, $A_c$ the triangle area, and $t_0$ the fabric thickness. The normal load
carried by a patch of fabric, $\int (E_t/t_0)\,(h-d)\,\mathrm{d}A$, is then
independent of the triangulation, and mesh refinement does not change the grasp
load. Triangles that transiently cross the shell boundary by more than a threshold
are projected back to the shell after the position update, with forces evaluated at
the pre-projection positions, so the projection resolves only deep transients that
the force law did not stop and does not alter the grasp force.

\paragraph{Friction and release.} The tangential term is anchored Coulomb friction
with a true static regime, following the classical elastoplastic
formulation~\citep{laursen1993,wriggers2006}. The anchor is stored in rigid-local
coordinates; $\mathbf{p}_c$ denotes its current world position and $\mathbf{x}_c$ the
cloth material point (Appendix~\ref{app:sim-contact}). With $\mathbf{P}_c$ the tangent
projector, the trial force
$\mathbf{f}_t = -k_t\mathbf{P}_c(\mathbf{x}_c - \mathbf{p}_c)$ is limited to the
Coulomb cone $\|\mathbf{f}_t\| \le \mu f_n$, and when the limit is exceeded the
force is clamped and the anchor is updated by return mapping, giving sliding;
$k_t = \kappa k_c$ places the pre-sliding displacement
at the scale of fabric. What the static regime buys is visible by contrast with
released stacks, which replace the stick regime by a velocity-regularised
ramp~\citep{li2020ipc}: a contact loaded to a fraction $\rho = \|\mathbf{f}_t\| /
(\mu f_n)$ of the cone then creeps at the steady velocity
\begin{equation}
v = \varepsilon\,\big(1 - \sqrt{1-\rho}\,\big),
\label{eq:creep}
\end{equation}
with $\varepsilon$ the regularisation velocity, non-zero for any $\rho > 0$;
holding fabric against gravity then requires an inflated $\mu$, and an inflated
$\mu$ locks fabric wrapped around a finger tip on release through the capstan factor
$e^{\mu\theta}$. With the anchor, $v = 0$ for $\rho < 1$, holding depends on
$\mu f_n$ alone without inflating $\mu$ (one cloth-side coefficient serves all
rigid surfaces, Appendix~\ref{app:sim-contact}), and when the gripper opens $f_n$
and hence $\mathbf{f}_t$ vanish and the fabric leaves the finger under its own
weight.

Grasp, hold, and release thus arise from the forces of Eq.~(\ref{eq:contact})
alone: no attachment constraints, no collision disabling on release, no adhesion
term, and no filtering of failed trajectories.

\subsection{Co-Simulation with the Rigid-Body Solver}
\label{sec:cosim}

\paragraph{Bidirectional force coupling.} Rigid motion drives the cloth through
Eq.~(\ref{eq:contact}) at the fingers and through vertex contact at free objects; the same forces act on the rigid bodies with equal magnitude
and opposite direction. After the cloth solve of a substep converges, the normal and
tangential force of every contact is evaluated at the converged positions and
accumulated per rigid body $b$ into a force $\mathbf{F}_b = -\sum_{c\in b}
\mathbf{f}_c$ and a torque about its centre of mass $\boldsymbol{\tau}_b =
-\sum_{c\in b}(\mathbf{x}_c - \mathbf{c}_b)\times\mathbf{f}_c$, applied by the
rigid-body solver~\citep{todorov2012mujoco} as an external load in the next substep.
Force transmission is therefore bidirectional: a finger feels the load of the fabric
it holds, a packed item is supported by the bag bottom, and a lifted bag carries the
weight of its contents to the gripper. The coupling is a staggered scheme at substep
granularity.

\paragraph{Gripper actuation.} The gripper is a PD position actuator with a force
limit. Closing on fabric saturates the actuator, so the grip force equals the
actuator's stall force and is independent of the residual jaw opening: a settable
physical quantity rather than a stiffness times a leftover gap.

\paragraph{Stability of the staggered scheme.} The contact force computed at the
start of a substep is held over that substep, so the rigid-body integrator sees it
as an explicit spring. For one finger degree of freedom the total contact stiffness
it can be exposed to is bounded by
\begin{equation}
k_{\mathrm{crit}} = \frac{4M}{\Delta t^{2}} + \frac{2c}{\Delta t} - k_p,
\label{eq:kcrit}
\end{equation}
with $M$ the finger mass including armature, $c$ its damping including the velocity
gain, and $k_p$ the position gain, the same form as the passivity bound of haptic
virtual walls~\citep{colgate1997}. For our gripper the summed stiffness of
Eq.~(\ref{eq:contact}) over a grasp lies a factor of 3--6 below this
bound (Appendix~\ref{app:sim-cosim}), whereas the much stiffer response of fully compacted
fabric would exceed it; the normal law is therefore linear, and a nonlinear
compression law is left to future work on fully implicit coupling.

\section{Benchmark Design}
\label{sec:benchmark}

\paragraph{Platform.} We use a dual-arm Piper platform with a 14-dimensional state
and action interface: six joints and one gripper value per arm. Policies receive RGB
observations from one overhead and two wrist cameras at 25~Hz. Simulation couples
our cloth solver with MuJoCo rigid-body dynamics and uses Omniverse Kit for rendering.
Simulated camera poses and intrinsics follow the real setup; detailed camera and data
specifications are provided in Appendix~\ref{app:data-interface}.

\subsection{Task Design}
\label{sec:task-design}
\begin{figure}[t]
\centering
\begingroup
\newcommand{\taskpanel}[6]{%
  \begin{minipage}[t]{0.485\textwidth}
  \centering
  \scriptsize\bfseries #1\par\vspace{1pt}
  \setlength{\tabcolsep}{0.4pt}%
  \begin{tabular}{@{}ccccc@{}}
  #2 & #3 & #4 &
  #5 & #6
  \end{tabular}
  \end{minipage}}
\setlength{\tabcolsep}{2pt}
\begin{tabular}{@{}cc@{}}
\taskpanel{(a) Fold T-shirt}
{\includegraphics[width=0.194\linewidth]{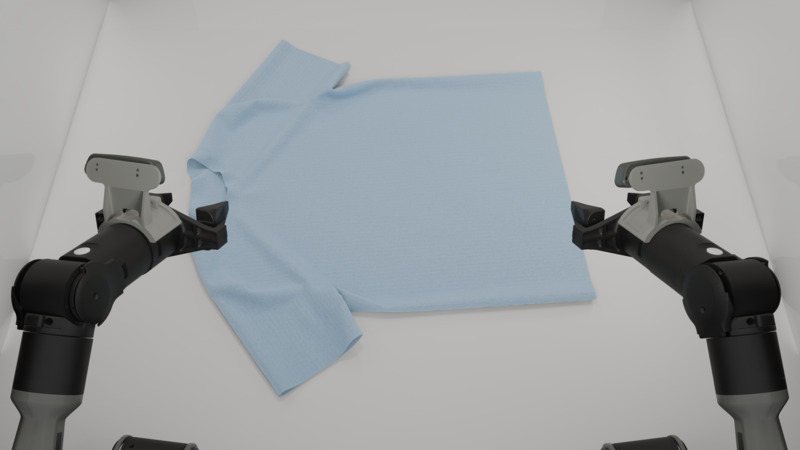}}
{\includegraphics[width=0.194\linewidth]{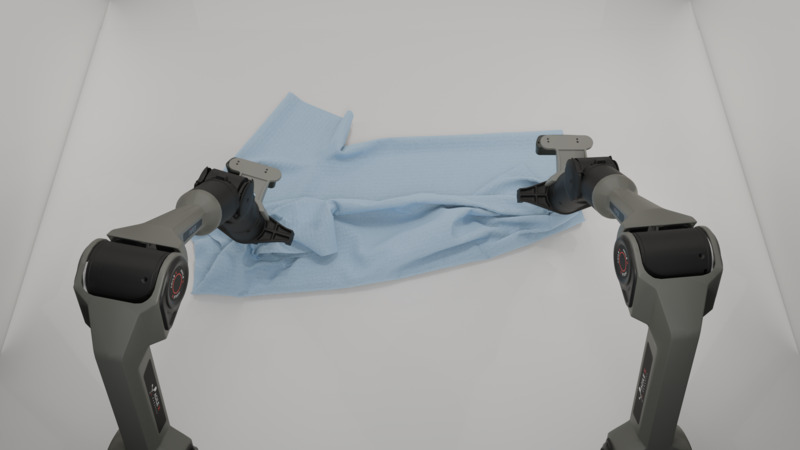}}
{\includegraphics[width=0.194\linewidth]{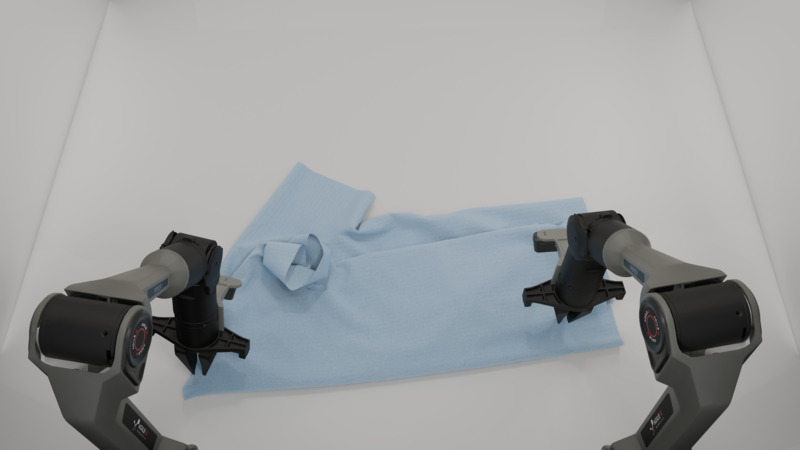}}
{\includegraphics[width=0.194\linewidth]{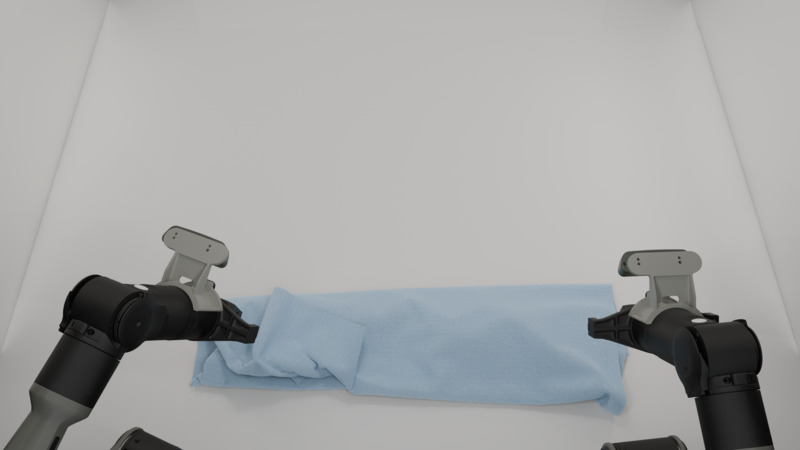}}
{\includegraphics[width=0.194\linewidth]{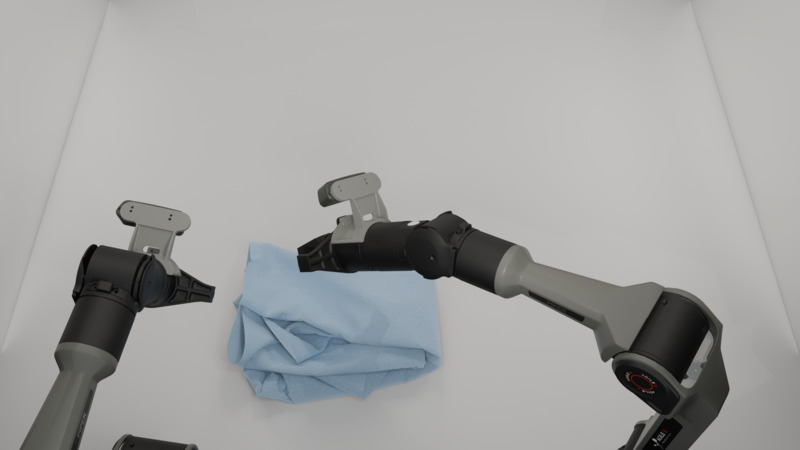}}
&
\taskpanel{(b) Flatten T-shirt}
{\includegraphics[width=0.194\linewidth]{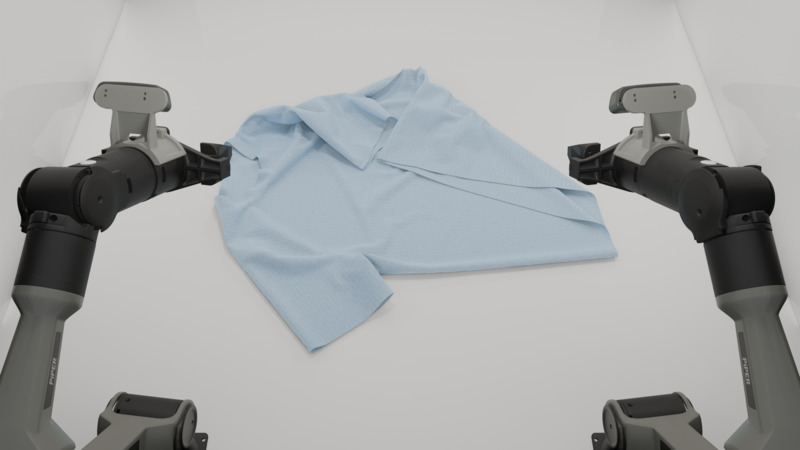}}
{\includegraphics[width=0.194\linewidth]{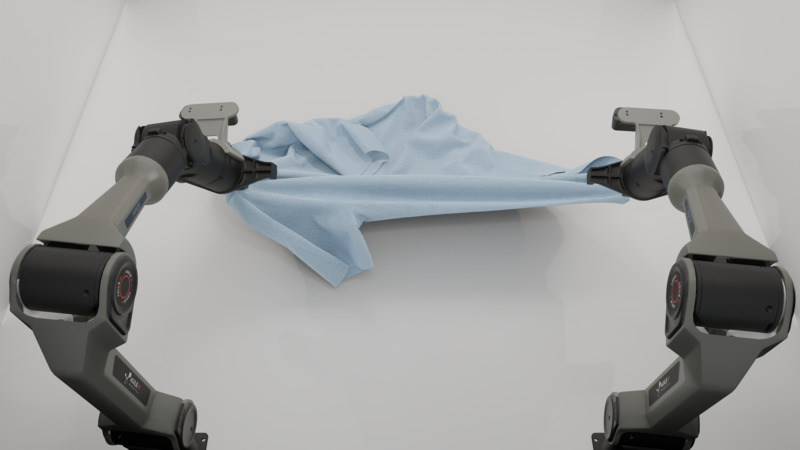}}
{\includegraphics[width=0.194\linewidth]{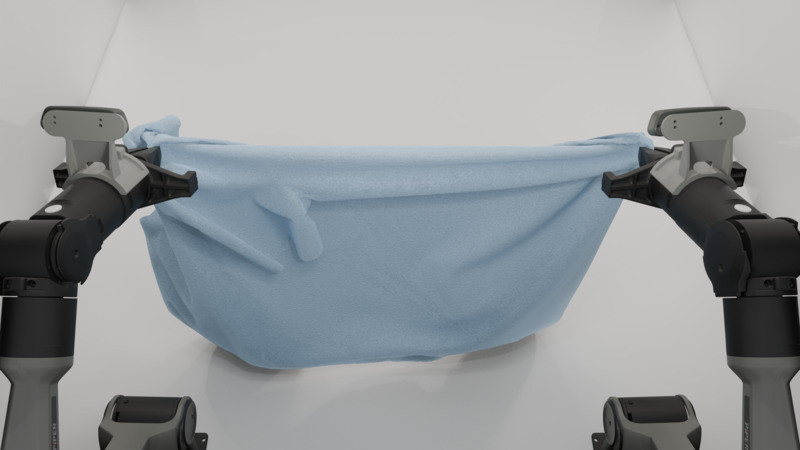}}
{\includegraphics[width=0.194\linewidth]{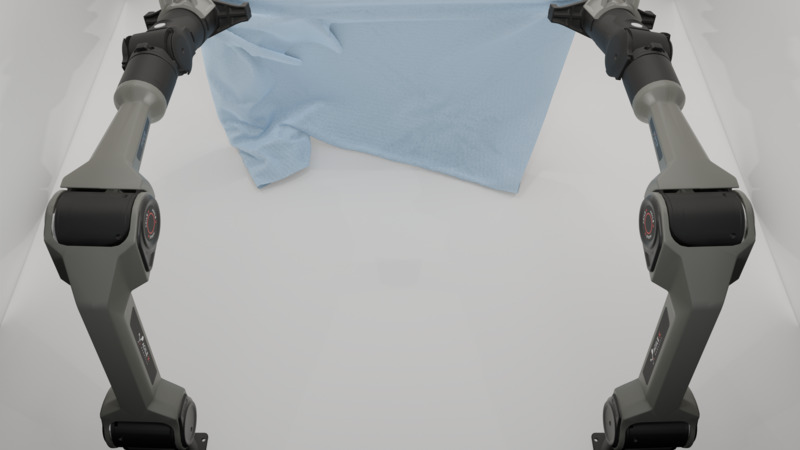}}
{\includegraphics[width=0.194\linewidth]{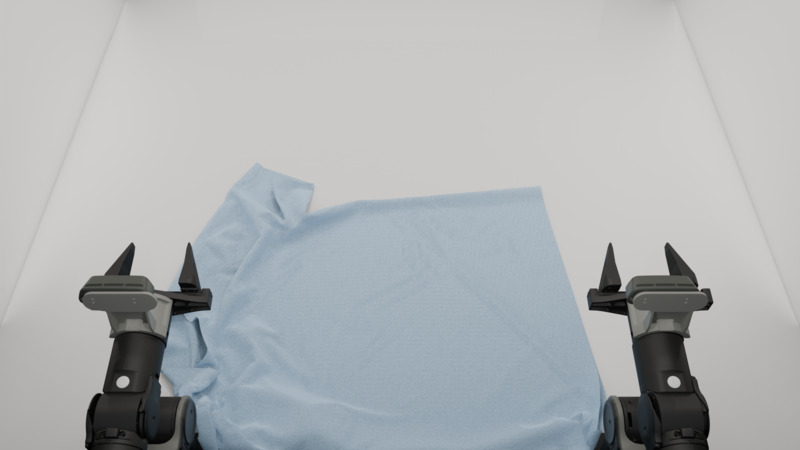}}
\\[5pt]
\taskpanel{(c) Flatten silk}
{\includegraphics[width=0.194\linewidth]{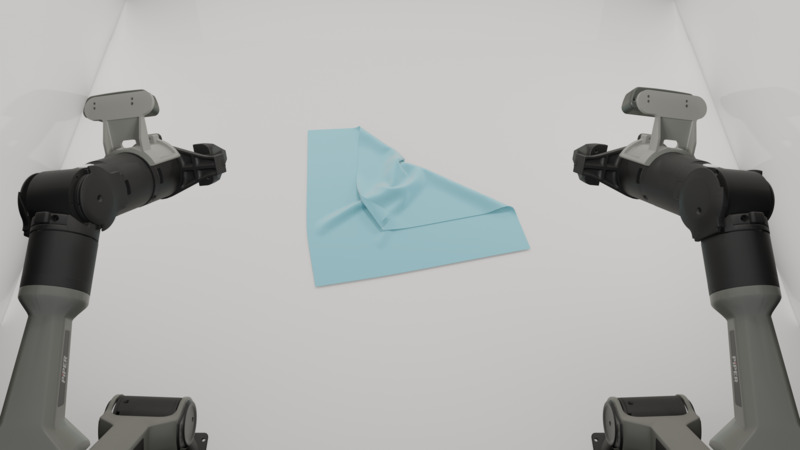}}
{\includegraphics[width=0.194\linewidth]{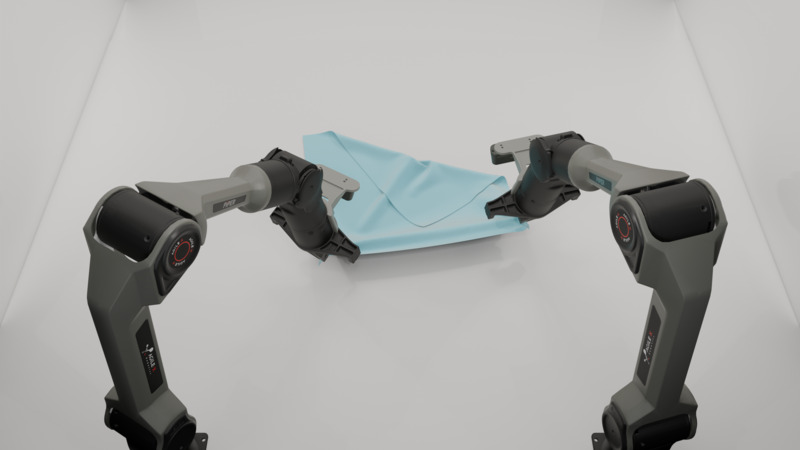}}
{\includegraphics[width=0.194\linewidth]{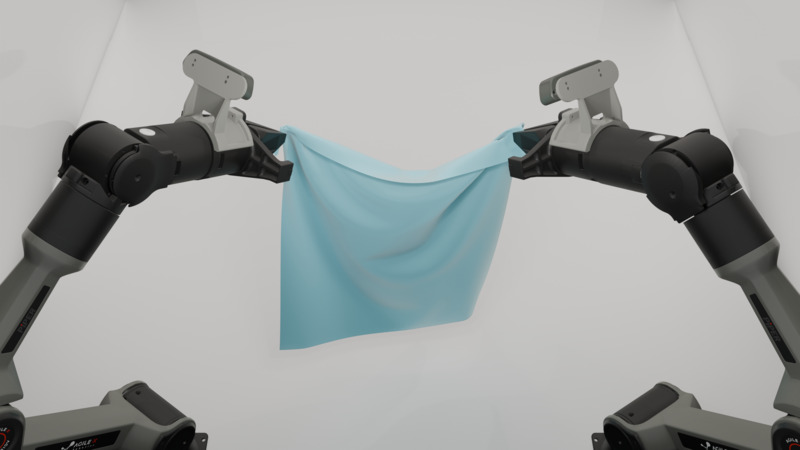}}
{\includegraphics[width=0.194\linewidth]{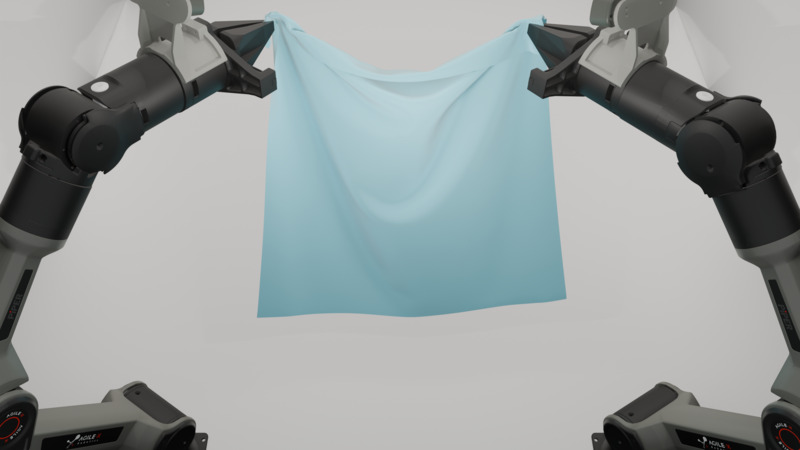}}
{\includegraphics[width=0.194\linewidth]{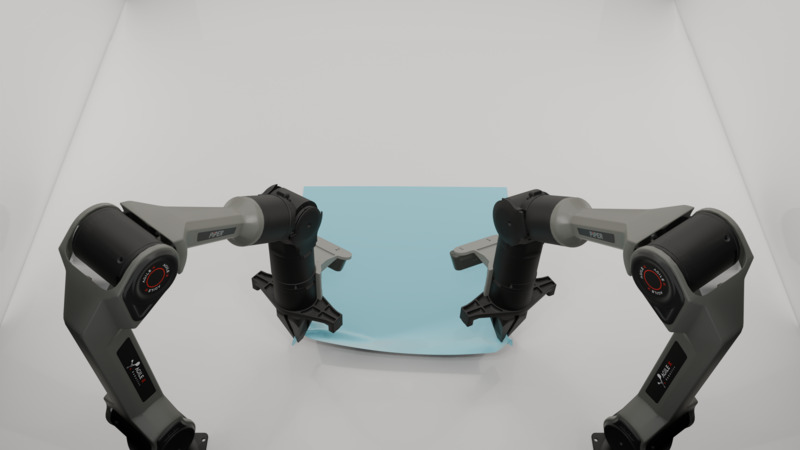}}
&
\taskpanel{(d) Lift silk}
{\includegraphics[width=0.194\linewidth]{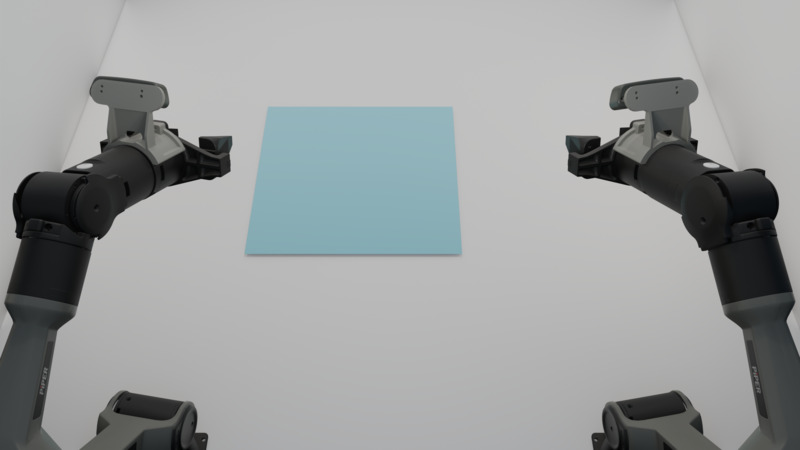}}
{\includegraphics[width=0.194\linewidth]{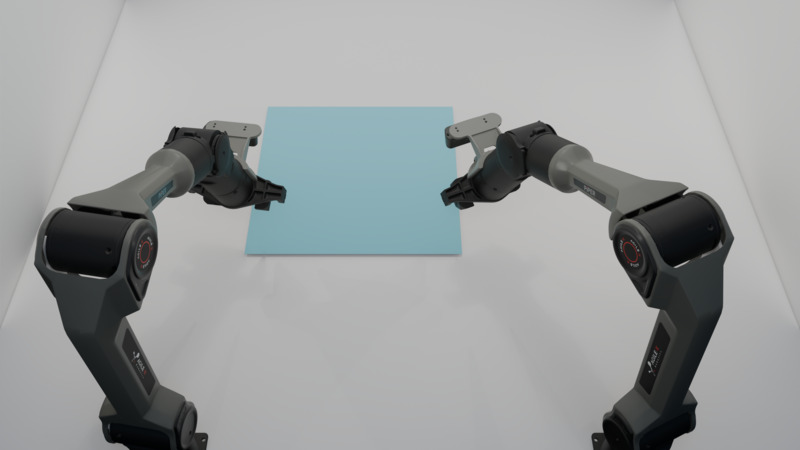}}
{\includegraphics[width=0.194\linewidth]{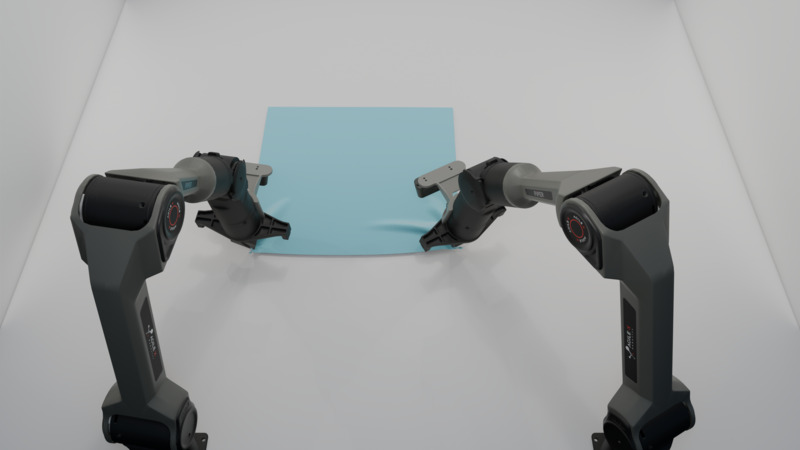}}
{\includegraphics[width=0.194\linewidth]{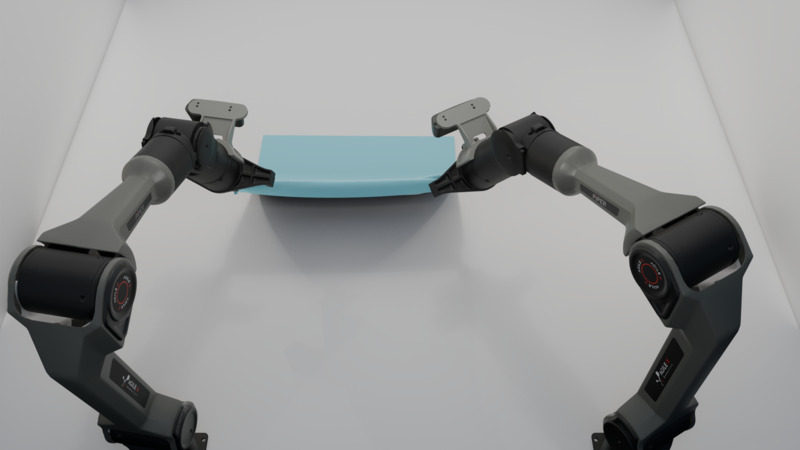}}
{\includegraphics[width=0.194\linewidth]{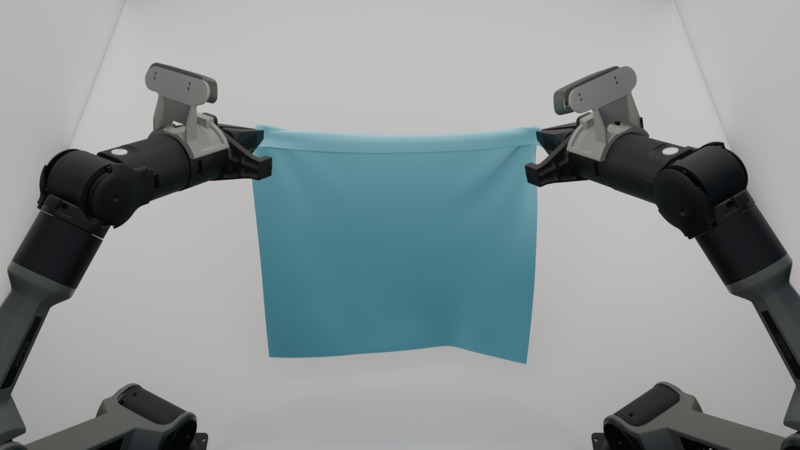}}
\\[5pt]
\taskpanel{(e) Carry bag}
{\includegraphics[width=0.194\linewidth]{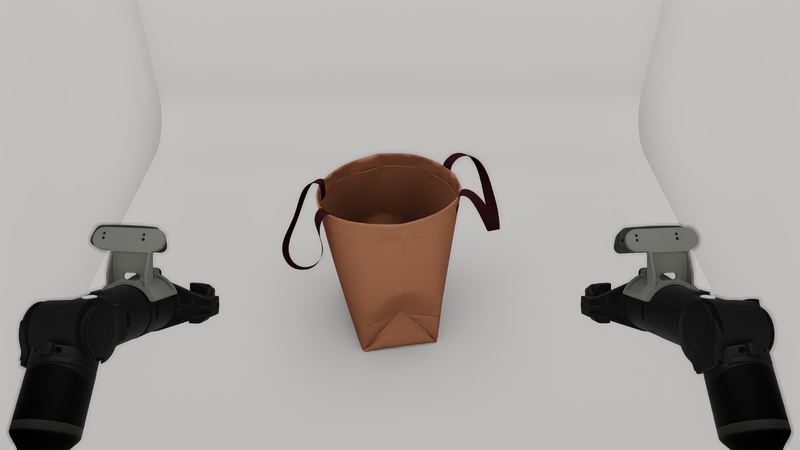}}
{\includegraphics[width=0.194\linewidth]{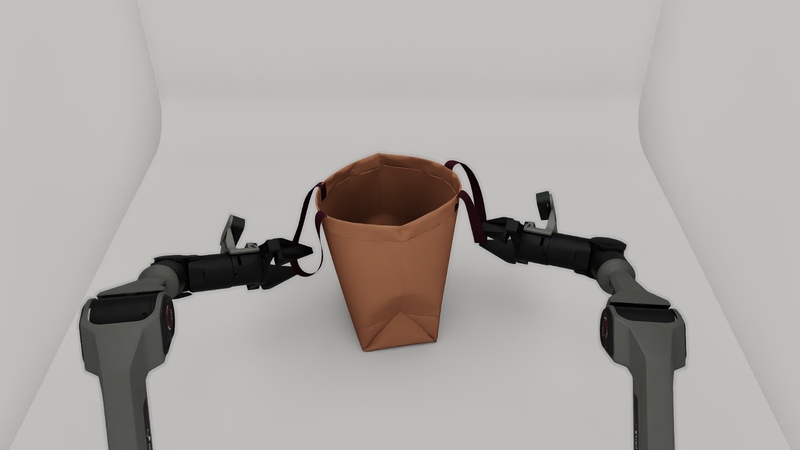}}
{\includegraphics[width=0.194\linewidth]{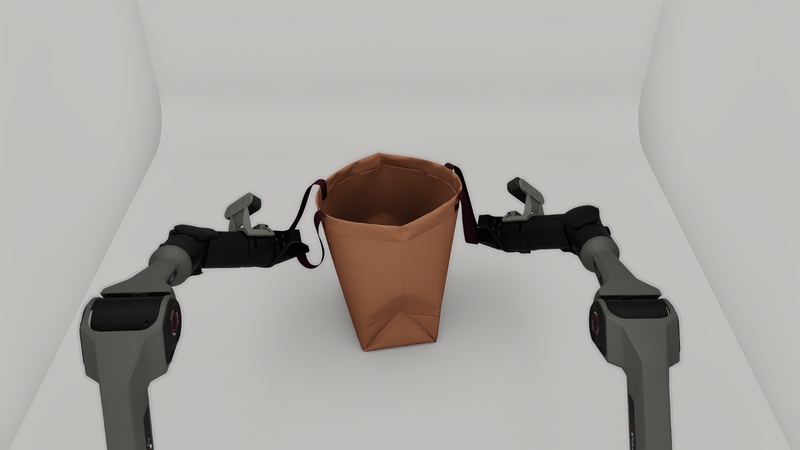}}
{\includegraphics[width=0.194\linewidth]{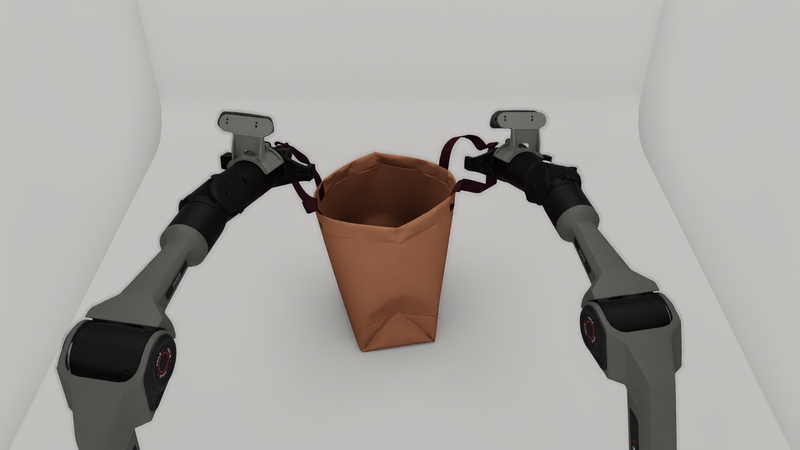}}
{\includegraphics[width=0.194\linewidth]{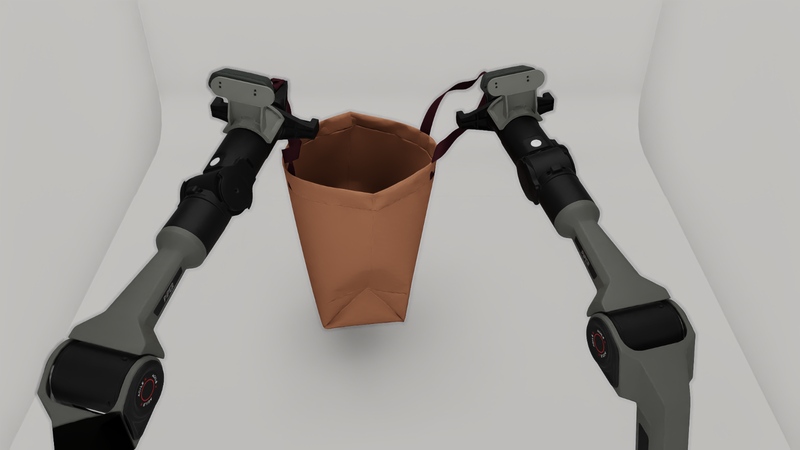}}
&
\taskpanel{(f) Pack snack}
{\includegraphics[width=0.194\linewidth]{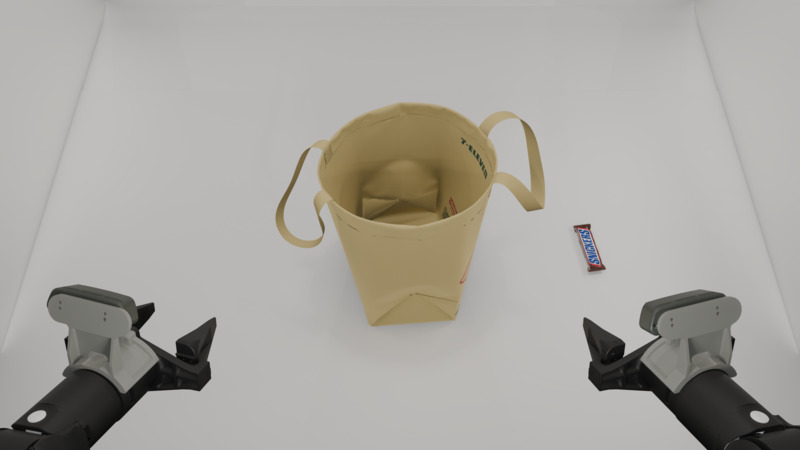}}
{\includegraphics[width=0.194\linewidth]{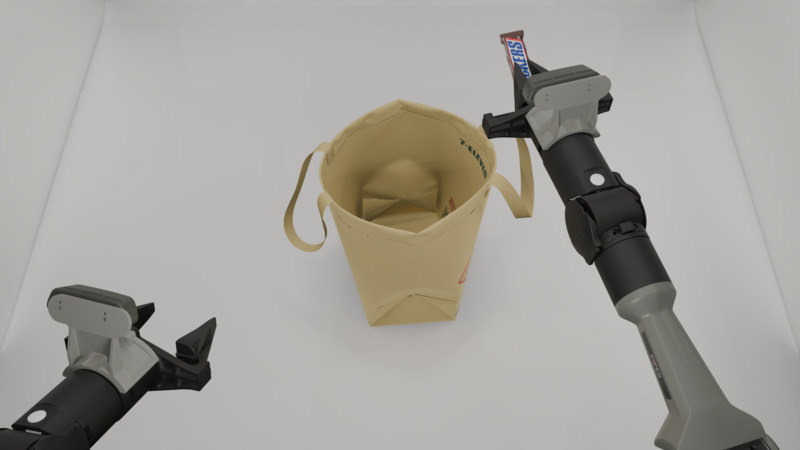}}
{\includegraphics[width=0.194\linewidth]{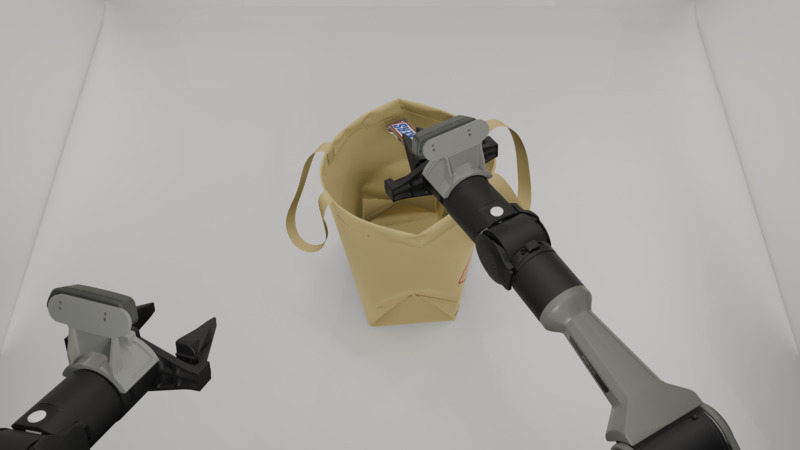}}
{\includegraphics[width=0.194\linewidth]{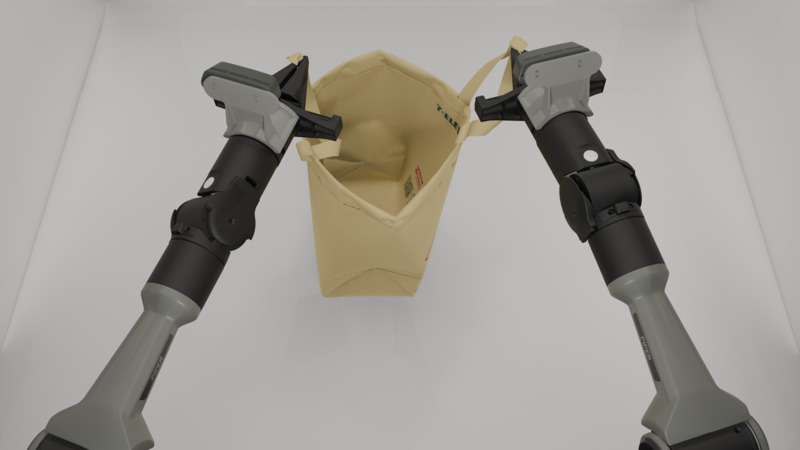}}
{\includegraphics[width=0.194\linewidth]{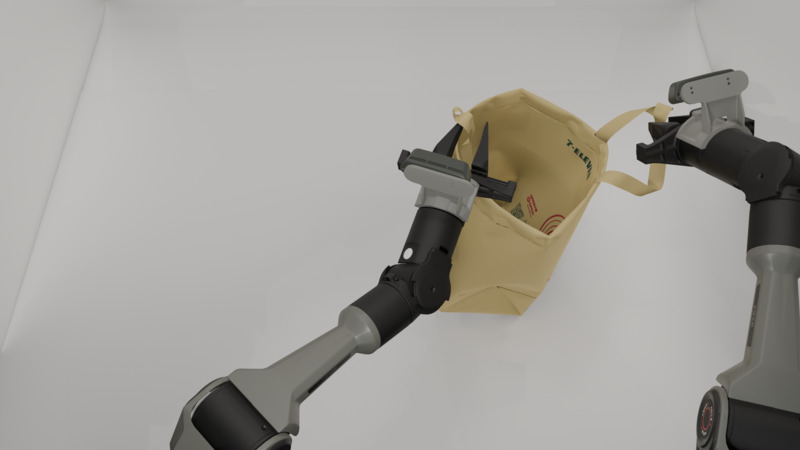}}
\end{tabular}
\endgroup
\caption{\textbf{The \benchname{} task suite.} Each panel shows five stages of one
successful episode, progressing from left to right. The suite covers (a) quasi-static,
long-horizon folding with self-contact; (b--c) flattening under occlusion for garments
with distinct material properties; (d) slip-sensitive, attachment-free bimanual
grasping; (e) lifting a volumetric bag by its handles; and (f) rigid-object
placement into a deformable container followed by loaded-bag transport. Exact success
criteria are given in Appendix~\ref{app:tasks}.}
\label{fig:tasks}
\end{figure}

We first summarize common deformable-manipulation situations with five physical
regimes: (i) quasi-static, long-horizon folding with self-contact; (ii) configuration
recovery under severe self-occlusion; (iii) dynamic momentum transfer during
flinging; (iv) friction-limited grasping and hold maintenance without attachment; and
(v) manipulation of a volumetric bag whose handles must carry its load, including
rigid contents when packed. These regimes are not mutually exclusive task categories: a single
task can exercise several of them. We then instantiate the coverage with six
representative tasks, shown in Figure~\ref{fig:tasks}. Fold T-shirt primarily tests
(i); Flatten T-shirt and Flatten silk combine (ii) and (iii) on garments with
different material response; Lift silk isolates (iv); Carry bag tests (v) through
bimanual handle lifting; and Pack snack combines (v) with rigid pick-and-place, containment,
and loaded-bag transport. Success criteria are outcome-based rather than tied to a
prescribed grasp point or trajectory, so a policy receives credit for any behavior
that reaches the target state.

\subsection{Assets}
\label{sec:assets}
Asset diversity is essential for a deformable benchmark, but meshes produced by generative
3D models~\citep{wang2025embodiedgen,hunyuan3d2025} cannot be used directly as simulated
cloth. They are typically double-layered closed surfaces rather than single-layer fabric,
are split into hundreds of disconnected pieces, and carry UV layouts packed as texture
atlases whose per-chart scale does not match the material. A cloth solver that reads the UV
layout as the rest pattern then starts from a distorted rest shape, and the garment becomes
unstable or diverges within the first frames (Appendix~\ref{app:asset-expansion}). We
therefore start from high-quality template meshes and expand them through a framework that
separates geometry, appearance, and physics: morphology variants edit semantic regions and
reconstruct a consistent rest pattern, appearance variants change only render materials and
printed patterns, and physical variants assign explicit material-parameter profiles, each
leaving the other factors unchanged. Every variant records its seed and recipe and can be
rebuilt from its source. We will release 10,000 expanded assets, each passing the same
simulation-stability check; the held-out appearance pools of
Section~\ref{sec:generalization} are drawn from them.

\subsection{Training Data}
\label{sec:training-data}
We generate 200 demonstration trajectories per task, recorded from three cameras (two
wrist cameras and one top camera) at 25~Hz (Appendix~\ref{app:data-interface}). Each task
is trained and evaluated on its own standard task distribution (Appendix~\ref{app:tasks}):
all tasks randomize object placement, and Carry bag additionally randomizes the handle
stiffness and the bag appearance. We train ACT, Diffusion Policy, the VLAs $\pi_{0.5}$ and
GR00T, and the WAMs LingBot-VA and Fast-WAM, each on the same data with its official configuration
(Appendix~\ref{app:policy-protocol}).

\subsection{Evaluation Protocol}
\label{sec:eval-protocol}

\paragraph{Evaluation axes.}
\label{sec:ladder}
Replacing a whole object changes its appearance, physical properties, and initial state
at once. For the two flattening tasks, we instead vary one factor at a time against the
same training condition, the C0 configuration with the canonical appearance. The
configuration ladder (C0--C2) changes only how the cloth is initially folded or dropped,
through seeded programs with matched asset, camera, and placement distribution
(Figure~\ref{fig:ladder}; generation details in Appendix~\ref{app:ladder-stats}). The
appearance axis assigns one of 50 held-out appearance variants (Section~\ref{sec:assets})
to each episode under C0, leaving geometry and physics unchanged.

\begin{figure}[t]
\centering
\begingroup
\setlength{\figfourimageheight}{0.080\textwidth}
\setlength{\figfourpanelheight}{0.145\textwidth}
\begin{minipage}[t][\figfourpanelheight][t]{0.465\linewidth}
\vspace{0pt}
\centering
\scriptsize\bfseries Configuration ladders\par\vspace{1pt}
\makebox[\linewidth][c]{\makebox[0.48\linewidth][c]{\scriptsize Silk}%
  \hfill\makebox[0.48\linewidth][c]{\scriptsize T-shirt}}\par\vspace{-1pt}
\begin{tabular}{@{}c@{\hspace{1pt}}c@{\hspace{1pt}}c@{\hspace{3pt}}c@{\hspace{1pt}}c@{\hspace{1pt}}c@{}}
\makebox[.168\linewidth][c]{\includegraphics[width=.168\linewidth,height=\figfourimageheight,keepaspectratio]{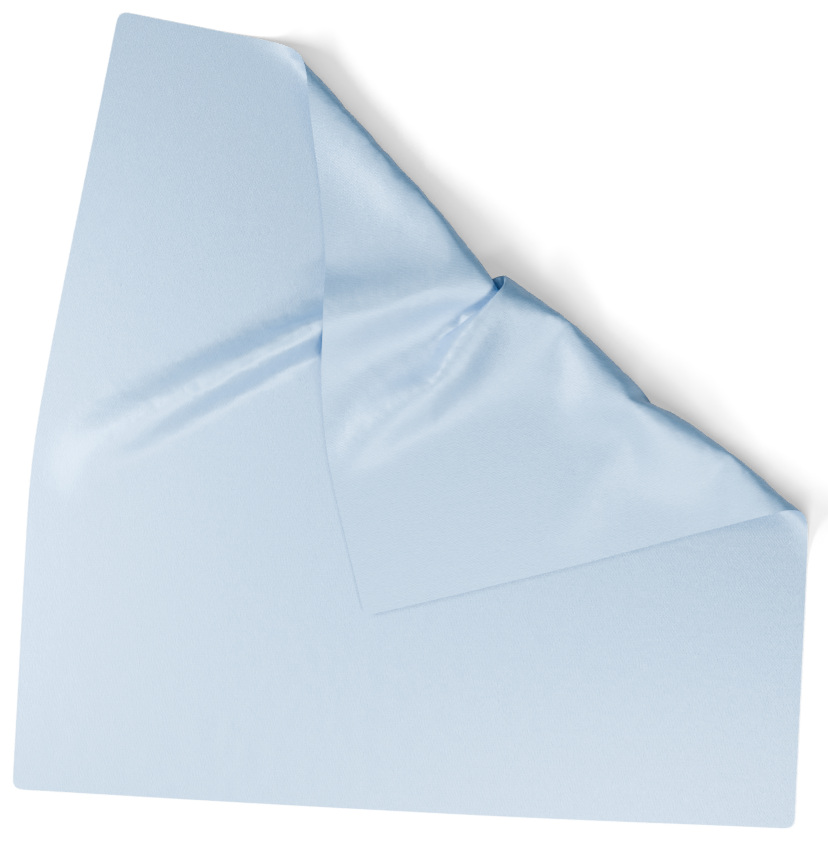}} &
\makebox[.122\linewidth][c]{\includegraphics[width=.122\linewidth,height=\figfourimageheight,keepaspectratio]{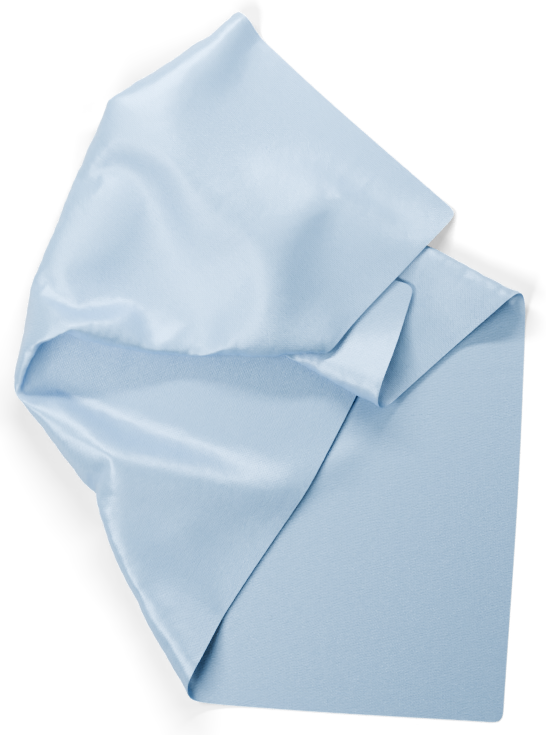}} &
\makebox[.163\linewidth][c]{\includegraphics[width=.163\linewidth,height=\figfourimageheight,keepaspectratio]{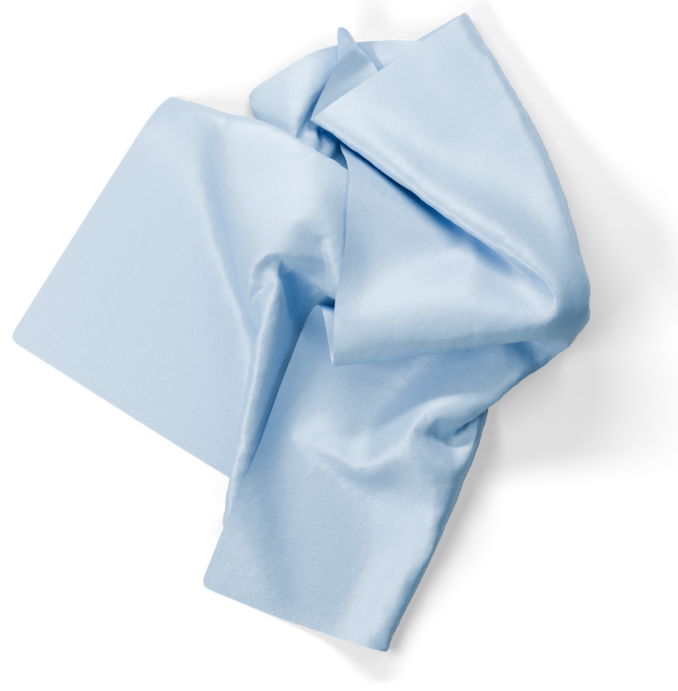}} &
\makebox[.190\linewidth][c]{\includegraphics[width=.190\linewidth,height=\figfourimageheight,keepaspectratio]{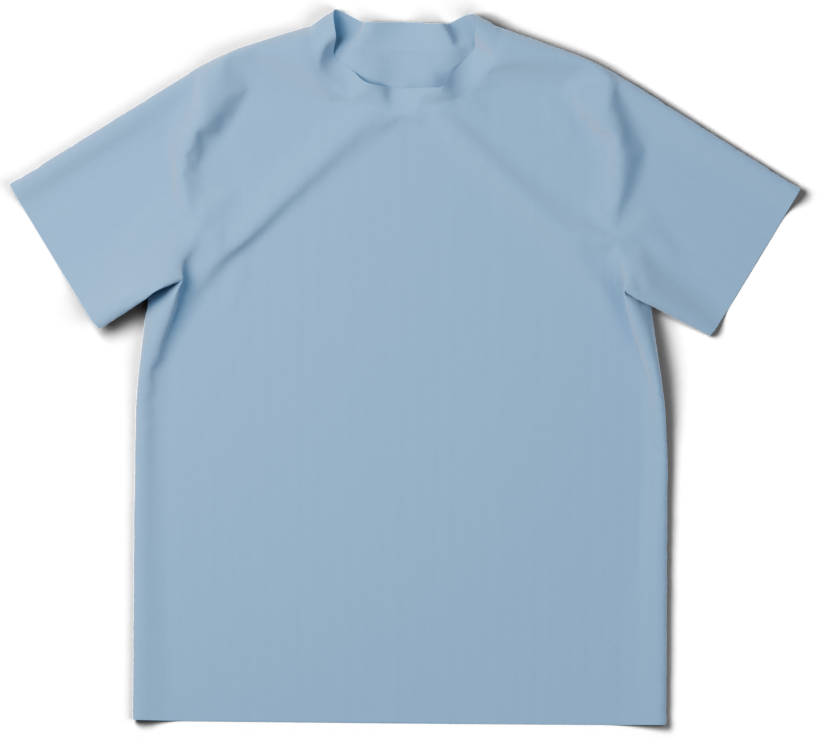}} &
\makebox[.156\linewidth][c]{\includegraphics[width=.156\linewidth,height=\figfourimageheight,keepaspectratio]{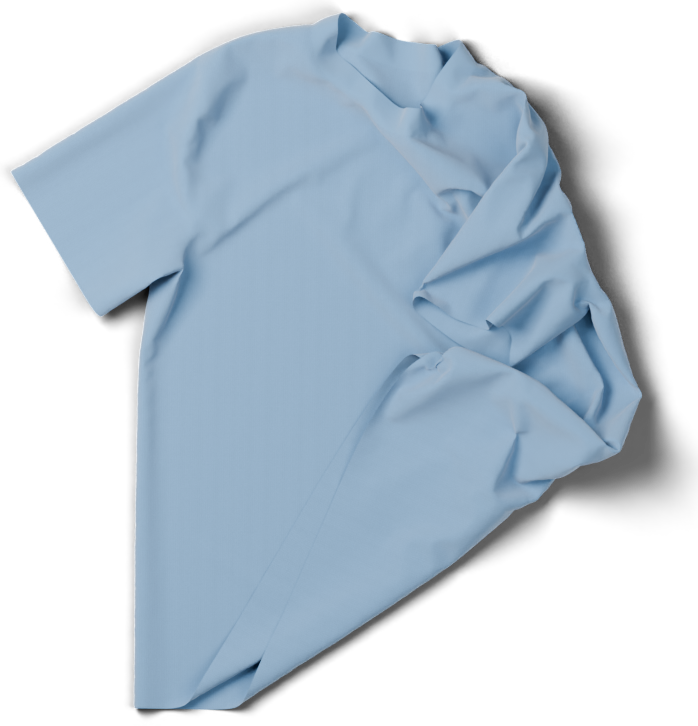}} &
\makebox[.163\linewidth][c]{\includegraphics[width=.163\linewidth,height=\figfourimageheight,keepaspectratio]{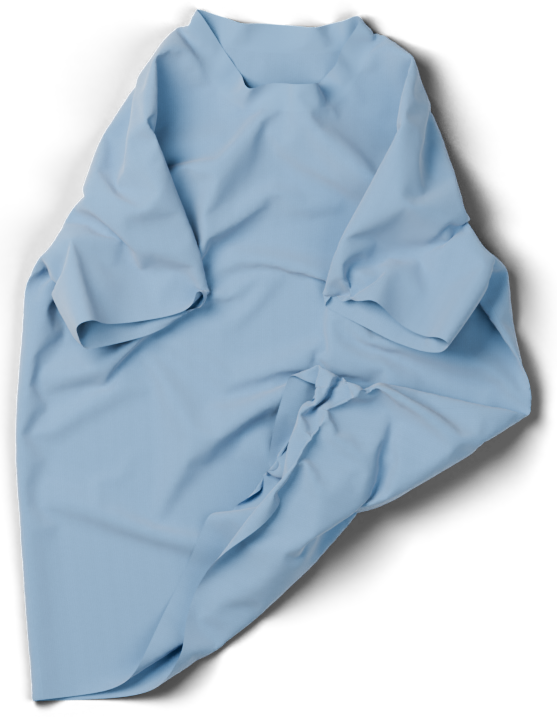}}
\end{tabular}
\vfill
\begin{tabular}{@{}c@{\hspace{1pt}}c@{\hspace{1pt}}c@{\hspace{3pt}}c@{\hspace{1pt}}c@{\hspace{1pt}}c@{}}
\makebox[0.168\linewidth][c]{\scriptsize C0} &
\makebox[0.122\linewidth][c]{\scriptsize C1} &
\makebox[0.163\linewidth][c]{\scriptsize C2} &
\makebox[0.190\linewidth][c]{\scriptsize C0} &
\makebox[0.156\linewidth][c]{\scriptsize C1} &
\makebox[0.163\linewidth][c]{\scriptsize C2}
\end{tabular}
\end{minipage}\hfill%
\begin{minipage}[t][\figfourpanelheight][c]{0.012\linewidth}
\centering\rule{0.4pt}{\dimexpr\figfourpanelheight-2pt\relax}
\end{minipage}\hfill%
\begin{minipage}[t][\figfourpanelheight][t]{0.505\linewidth}
\vspace{0pt}
\centering
\scriptsize\bfseries Appearance variants\par\vspace{1pt}
\makebox[\linewidth][c]{\makebox[0.49\linewidth][c]{\scriptsize Silk}%
  \hfill\makebox[0.49\linewidth][c]{\scriptsize T-shirt}}\\[-1pt]
\makebox[0.49\linewidth][c]{%
  \includegraphics[height=\figfourimageheight]{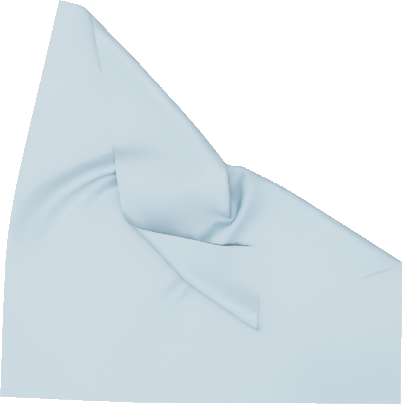}\hspace{-0.110\linewidth}%
  \includegraphics[height=\figfourimageheight]{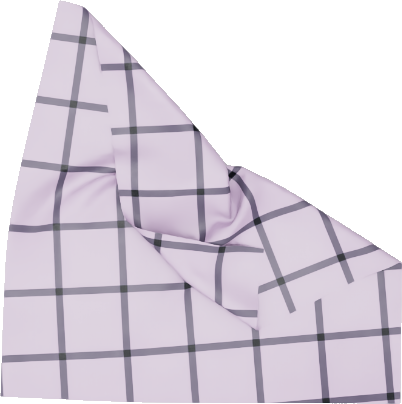}\hspace{-0.110\linewidth}%
  \includegraphics[height=\figfourimageheight]{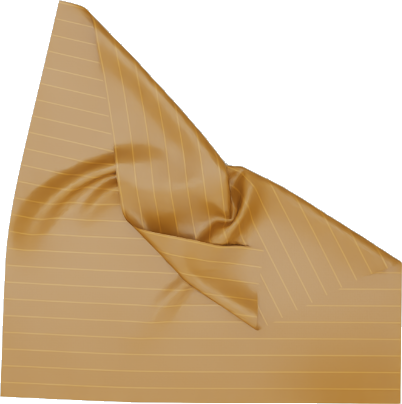}\hspace{-0.110\linewidth}%
  \includegraphics[height=\figfourimageheight]{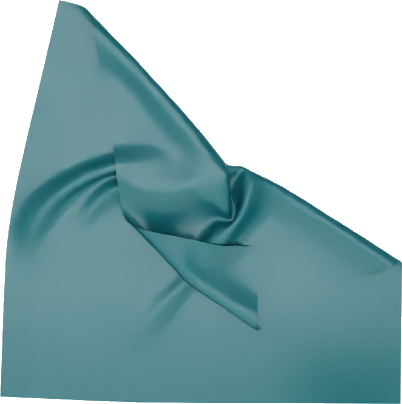}\hspace{-0.110\linewidth}%
  \includegraphics[height=\figfourimageheight]{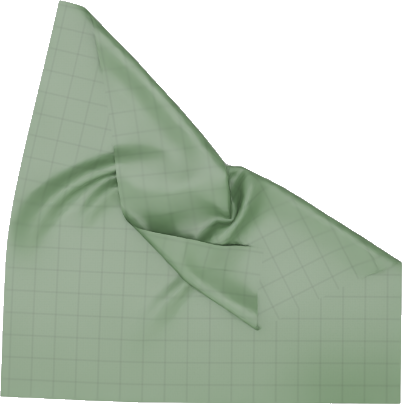}\hspace{-0.110\linewidth}%
  \includegraphics[height=\figfourimageheight]{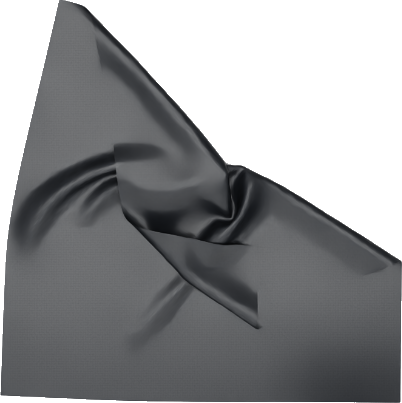}}\hfill%
\makebox[0.49\linewidth][c]{%
  \includegraphics[height=\figfourimageheight]{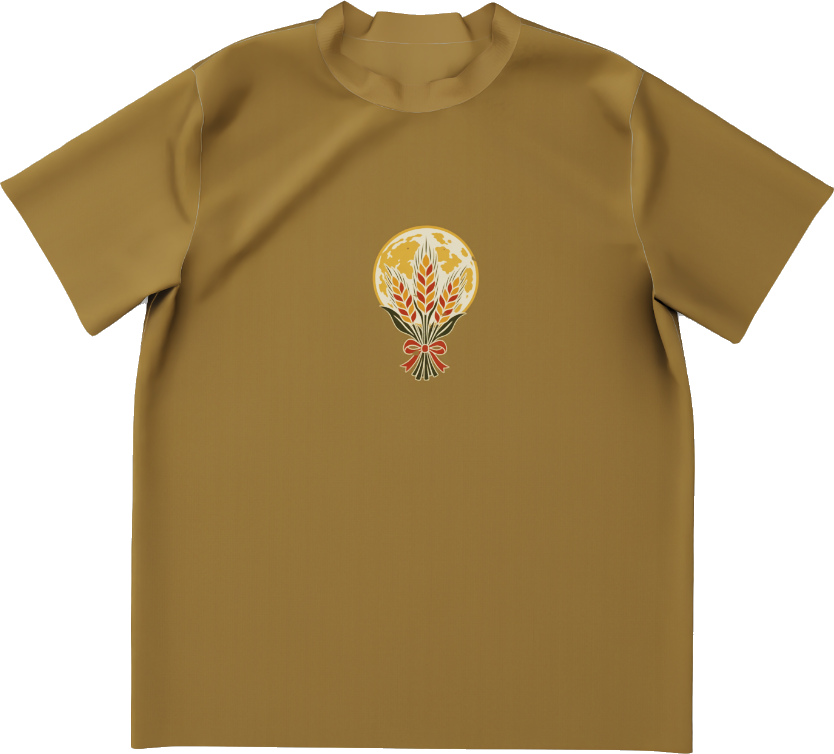}\hspace{-0.110\linewidth}%
  \includegraphics[height=\figfourimageheight]{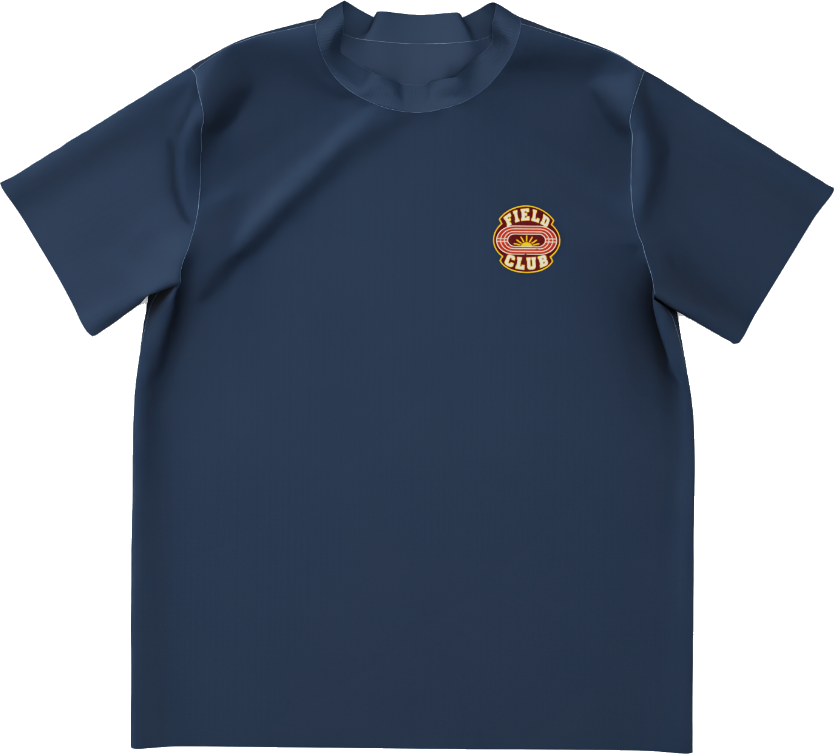}\hspace{-0.110\linewidth}%
  \includegraphics[height=\figfourimageheight]{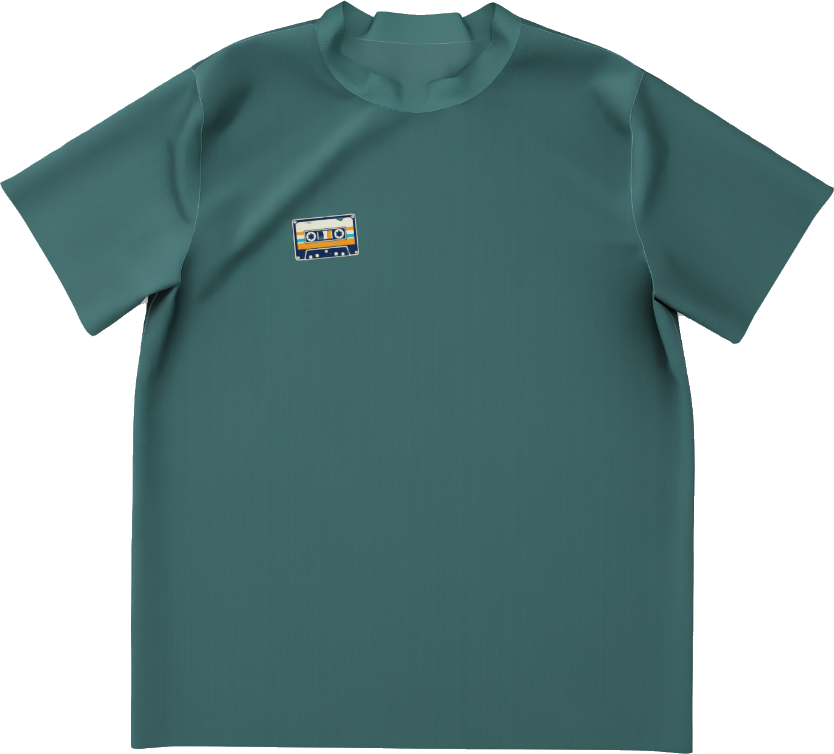}\hspace{-0.110\linewidth}%
  \includegraphics[height=\figfourimageheight]{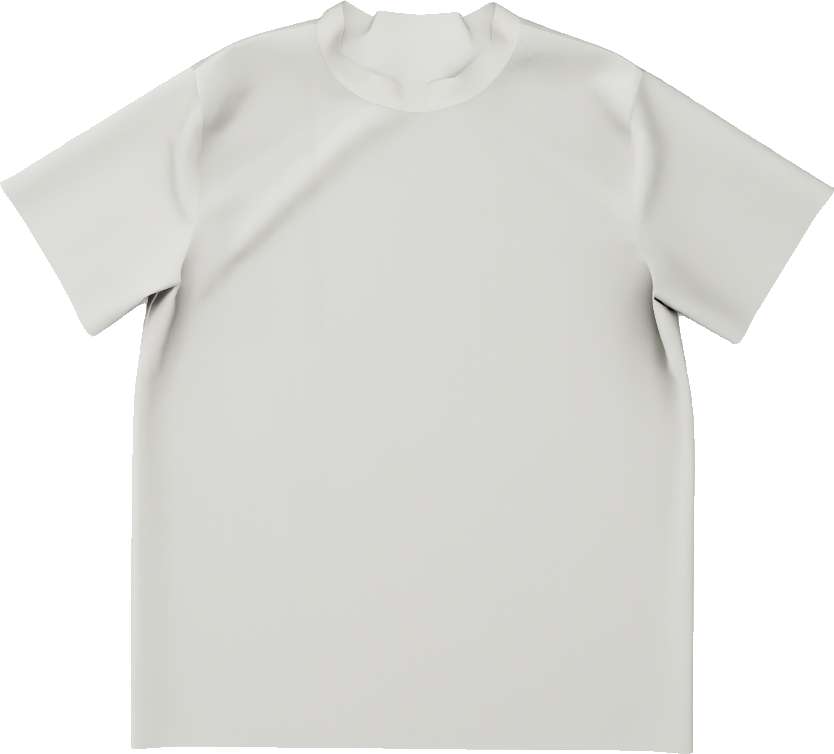}\hspace{-0.110\linewidth}%
  \includegraphics[height=\figfourimageheight]{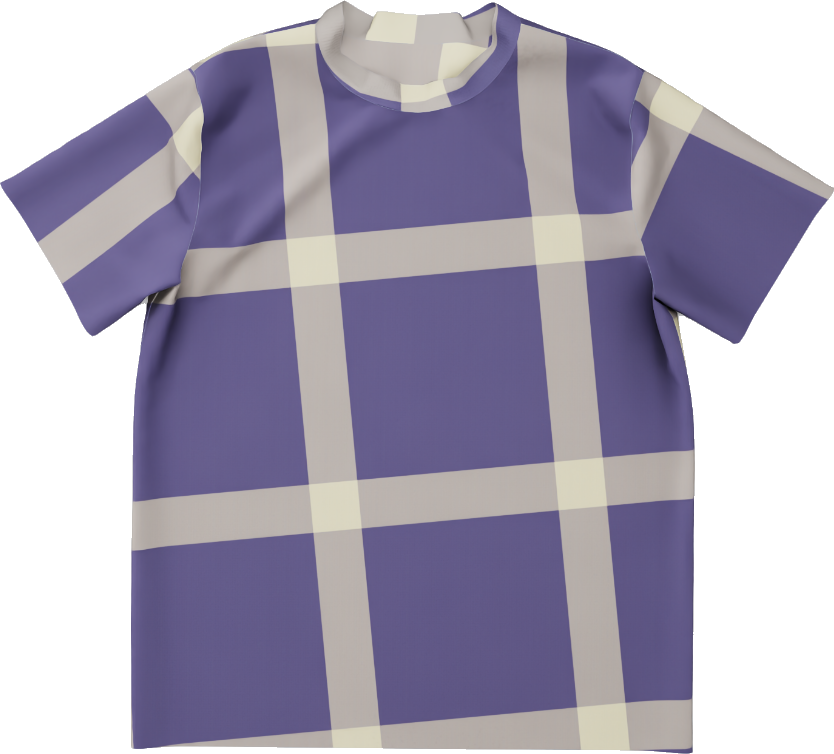}\hspace{-0.110\linewidth}%
  \includegraphics[height=\figfourimageheight]{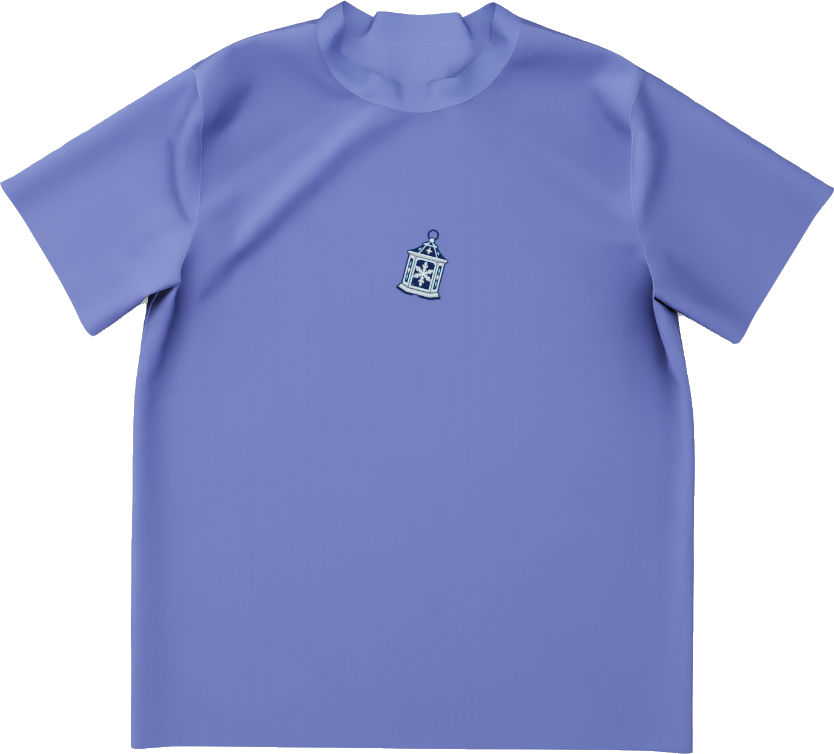}}\par
\vfill
\makebox[\linewidth][c]{\makebox[0.49\linewidth][c]{\scriptsize C0 init, V1--V6}%
  \hfill\makebox[0.49\linewidth][c]{\scriptsize V1--V6}}
\end{minipage}
\endgroup
\caption{\textbf{The two evaluation axes.} Left: configuration ladders. Silk: C0 folds one
far corner, C1 randomly folds one or two corners inward, C2 drops the cloth from a two-pin
hold. T-shirt: C0 folds the right sleeve corner, C1 folds the right half, C2 applies
random deep folds at the four corners. Right: appearance variants of the same assets
(silk shown after C0 initialization).}
\label{fig:ladder}
\end{figure}

\paragraph{Metrics.}
Each task reports an outcome-based binary success rate (SR), credited for any trajectory
that reaches the goal, and a continuous 0--100 score of partial progress
(Appendix~\ref{app:tasks}). For a shifted condition $x$, we report the absolute change
$\Delta_x=S_x-S_{\mathrm{C0}}$ and, when $S_{\mathrm{C0}}>0$, the normalized degradation
$D_x=(S_{\mathrm{C0}}-S_x)/S_{\mathrm{C0}}$.

\paragraph{Protocol.}
Each cell uses 50 episodes from a fixed evaluation seed schedule, disjoint from
collection and shared by all policies together with the same episode-to-condition
manifest. Evaluation uses no augmentation, and no held-out episode is used for
training, checkpoint selection, or tuning.

\section{Experiments}

\subsection{Simulator Validation}
\label{sec:validation}

We examine whether grasp, hold, and release arise from contact forces alone
(Table~\ref{tab:xsim}) and whether each contact component is necessary
(Table~\ref{tab:ablation}).

\paragraph{Cross-simulator comparison.} A recorded dual-arm grasp on a flat T-shirt is
replayed at 15 placements, three times each, using the simulator configurations
detailed in Appendix~\ref{app:sim-criteria}. Isaac Sim's PBD cloth forms
no bimanual grasp. Released Style3D and Newton VBD hold but never release, as the
fingers stay in the fabric; VBD also diverges in 12 of 45 trials. Only our
contact grasps, holds, and releases in every trial, still at a lower step cost than Isaac
Sim and VBD.

\rev{\paragraph{Physical fidelity.} On the real robot, this trajectory grasps, holds, and releases
the T-shirt, and only our contact reproduces all three stages (Figure~\ref{fig:xsim-real}). On the
RGBench sim-to-real benchmark~\citep{hu2026rgbench}, our Chamfer distance to real point clouds is
0.054, versus 0.088 for Isaac Sim PBD and 0.128 for PyBullet (Appendix~\ref{app:rgbench}).}

\begin{table}[!t]
\small
\caption{\textbf{Cross-simulator grasp validation.} The same recorded dual-arm grasp,
replayed at 15 placements three times each (45 trials per simulator); success requires
grasp, hold, and release. Step time: wall-clock time per physical step on one RTX~4090
with a 9k-vertex T-shirt (Appendix~\ref{app:sim-criteria}). \rev{The trajectory reliably grasps, holds, and releases the T-shirt on the real robot
(Figure~\ref{fig:xsim-real}).}}
\label{tab:xsim}
\setlength{\tabcolsep}{3pt}
\begin{tabular*}{\linewidth}{@{\extracolsep{\fill}}l cccccc@{}}
\toprule
Simulator & Success $\uparrow$ & Grasp $\uparrow$ & Hold $\uparrow$ & Release $\uparrow$ & Diverged $\downarrow$ & ms / step $\downarrow$ \\
\midrule
Isaac Sim, PBD cloth               & 0/45  & 3/45  & 0/45  & --    & 0/45  & 5.3 \\
Newton VBD                         & 0/45  & 39/45 & 39/45 & 0/45  & 12/45 & 5.6 \\
Newton Style3D                     & 0/45  & 45/45 & 45/45 & 0/45  & 0/45  & 3.7 \\
Newton Style3D, our contact        & 45/45 & 45/45 & 45/45 & 45/45 & 0/45  & 4.8 \\
\bottomrule
\end{tabular*}
\end{table}

\begin{table}[!t]
\small
\caption{\textbf{Contact ablations.} (a) One component removed at a time (20 seeds). (b) Diffusion Policy on Lift silk with attach--detach grasping vs.\ our contact (50 episodes).}
\label{tab:ablation}
\centering
\setlength{\tabcolsep}{1.5pt}
\begin{tabular*}{\linewidth}{@{\extracolsep{\fill}}l ccc @{\hspace{1em}} l c@{}}
\toprule
(a) Variant & T-shirt release $\uparrow$ & Silk penetration $\downarrow$ & Pack lift $\uparrow$
  & (b) Grasp mode & Silk SR $\uparrow$ \\
\cmidrule(r{0.6em}){1-4}\cmidrule{5-6}
Our contact                        & 20/20 & 0/20  & 20/20 & Our contact    & 19/50 \\
\quad w/o triangle-level detection & 7/20  & 20/20 & 6/20  & Attach--detach & 9/50 \\
\quad w/o bidirectional coupling         & 20/20 & 1/20  & 0/20  &                &      \\
\bottomrule
\end{tabular*}
\end{table}

\paragraph{Attachment under closed-loop control.} Under the threshold-triggered
attach--detach of existing cloth benchmarks, the same checkpoints succeed less often: Diffusion Policy on Lift silk falls from 19/50 to 9/50 (Table~\ref{tab:ablation}b),
as spurious detachments drop the fabric mid-grasp. These failures stem from the
grasp model rather than the policy, so attach--detach underestimates policy performance.

\paragraph{Component ablation.} Each component is removed in turn
(Table~\ref{tab:ablation}a; 20 seeds per cell; edge seeding in
Appendix~\ref{app:sim-ablation}). With vertex spheres of the same material stiffness in
place of triangle-level SDF detection, the T-shirt stays on the fingers after the jaws
open, the silk intersects the opened fingers, and the bag task diverges in 9 of 20 runs.
Without bidirectional coupling, the bag is lifted but the packed item stays on the table in
every run; a check without the load-following condition of Pack snack
(Appendix~\ref{app:tasks}) would still pass 18 of these runs. Triangle-level detection thus
governs release and non-penetration; bidirectional coupling governs load transfer.

\subsection{Benchmark Results}
Table~\ref{tab:main} reports the success rate and mean score of the six policies on the six
tasks, each evaluated on its standard distribution. The mean score awards stage-wise partial
credit and therefore distinguishes policies with the same success rate.

\begin{table}[!h]
\caption{\textbf{Benchmark results in simulation.} Success rate (\%) and mean score (in
parentheses) on each task's standard distribution (C0 for the flattening tasks);
$N{=}50$ per cell.}
\label{tab:main}
\centering
\footnotesize
\setlength{\tabcolsep}{2pt}
\begin{tabular*}{\linewidth}{@{\extracolsep{\fill}}l cccccc c@{}}
\toprule
Policy & Lift silk & Flatten silk & Flatten T-shirt & Carry bag & Pack snack & Fold T-shirt & Avg \\
\midrule
$\pi_{0.5}$ & \textbf{98}\scr{98.0} & \textbf{86}\scr{94.5} & 82\scr{82.4} & 66\scr{82.5} & 92\scr{97.0} & \textbf{36}\scr{63.6} & \textbf{77} \\
LingBot-VA & 66\scr{78.0} & \textbf{86}\scr{97.0} & 86\scr{83.0} & 50\scr{70.0} & \textbf{98}\scr{98.8} & \textbf{36}\scr{57.6} & 70 \\
Fast-WAM & 58\scr{78.0} & 64\scr{87.2} & \textbf{92}\scr{81.6} & \textbf{80}\scr{85.5} & 8\scr{12.4} & 0\scr{22.8} & 50 \\
GR00T & 70\scr{79.0} & 46\scr{80.4} & 72\scr{76.9} & 40\scr{58.5} & 54\scr{81.4} & 2\scr{24.2} & 47 \\
ACT & 28\scr{62.0} & 10\scr{70.3} & 70\scr{79.6} & 38\scr{62.5} & 52\scr{80.0} & 30\scr{54.4} & 38 \\
DP & 38\scr{44.0} & 0\scr{57.3} & 78\scr{78.8} & 8\scr{27.0} & 18\scr{53.2} & 0\scr{16.2} & 24 \\
\bottomrule
\end{tabular*}
\end{table}

\subsection{Generalization Beyond the Training Condition}
\label{sec:generalization}
Table~\ref{tab:shift} evaluates the two flattening tasks along the configuration ladder and
on the 50 held-out appearance variants, with training fixed to C0 and the canonical
appearance. Performance generally degrades under both the configuration and the appearance
shifts; Section~\ref{sec:findings} analyzes the two axes.

\begin{table}[!h]
\caption{\textbf{Generalization under configuration and appearance shift.} Success rate (\%)
and mean score (in parentheses); training on C0 with canonical appearance; $N{=}50$ per cell.
C1/C2 follow Figure~\ref{fig:ladder}; C0\,+\,app.\ keeps C0 and applies 50 held-out appearance variants.}
\label{tab:shift}
\centering
\footnotesize
\setlength{\tabcolsep}{1.5pt}
\begin{tabular*}{\linewidth}{@{\extracolsep{\fill}}l c@{\hspace{0.9em}}cc@{\hspace{0.9em}}c @{\hspace{1.4em}} c@{\hspace{0.9em}}cc@{\hspace{0.9em}}c@{}}
\toprule
 & \multicolumn{4}{c}{Flatten T-shirt} & \multicolumn{4}{c}{Flatten silk} \\
\cmidrule(r{0.8em}){2-5}\cmidrule(l{0.8em}){6-9}
Policy & C0 & C1 & C2 & C0 + app. & C0 & C1 & C2 & C0 + app. \\
\midrule
$\pi_{0.5}$ & 82\scr{82.4} & 44\scr{73.4} & 12\scr{35.7} & 76\scr{80.3} & \textbf{86}\scr{94.5} & 20\scr{60.4} & \textbf{2}\scr{42.2} & 64\scr{85.3} \\
LingBot-VA & 86\scr{83.0} & \textbf{62}\scr{73.5} & 10\scr{28.2} & \textbf{84}\scr{77.1} & \textbf{86}\scr{97.0} & \textbf{22}\scr{61.6} & 0\scr{40.4} & \textbf{86}\scr{97.0} \\
Fast-WAM & \textbf{92}\scr{81.6} & 34\scr{71.8} & \textbf{14}\scr{27.9} & 42\scr{64.1} & 64\scr{87.2} & 16\scr{70.7} & 0\scr{43.1} & 0\scr{74.8} \\
GR00T & 72\scr{76.9} & 34\scr{62.9} & \textbf{14}\scr{31.1} & 58\scr{69.2} & 46\scr{80.4} & 18\scr{71.0} & 0\scr{44.6} & 38\scr{77.8} \\
ACT & 70\scr{79.6} & 16\scr{36.1} & 6\scr{25.8} & 28\scr{59.1} & 10\scr{70.3} & 6\scr{60.5} & 0\scr{43.4} & 4\scr{65.5} \\
DP & 78\scr{78.8} & 32\scr{62.9} & 10\scr{26.5} & 50\scr{70.4} & 0\scr{57.3} & 0\scr{55.8} & 0\scr{41.7} & 0\scr{53.1} \\
\bottomrule
\end{tabular*}
\end{table}

\subsection{Sim-to-Real Consistency}
\label{sec:sim2real}
\begin{figure}[!b]
\centering
\includegraphics[width=\linewidth]{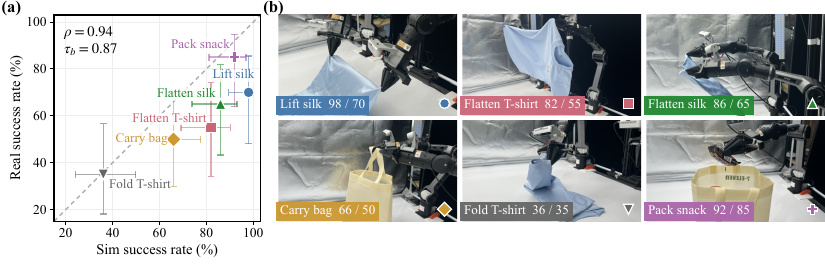}
\caption{\textbf{Sim-to-real evaluation of $\pi_{0.5}$.}
(a) Simulation ($N{=}50$)
vs.\ real-robot ($N{=}20$) success rate per task; whiskers are 95\% Wilson intervals,
dashed line is $y{=}x$. $\rho$ and $\tau_b$ measure agreement in task ordering, not in
absolute rate (bootstrap CIs in Appendix~\ref{app:sim2real}). (b) Real-robot frames;
colors and markers match (a).}
\label{fig:sim2real}
\end{figure}

Zero-shot sim-to-real transfer has been demonstrated for deformable
manipulation~\citep{simweaver2026}; here we ask a different question, namely whether the
difficulty ordering of tasks in simulation is preserved on the real robot. We deploy the $\pi_{0.5}$ checkpoints of Table~\ref{tab:main}, trained only on simulated
demonstrations, zero-shot on a real dual-arm robot and evaluate each task with 20 trials
using the same task-level success criteria as in simulation (Figure~\ref{fig:sim2real}). Across the six tasks,
simulated and real-robot success rates yield Spearman
$\rho=0.94$ and Kendall $\tau_b=0.87$. These results show that task
performance in simulation is strongly correlated with that on the real robot.
\rev{Without fine-tuning on \benchname{}, the pretrained $\pi_{0.5}$ succeeds in none of the
real trials on any task (Table~\ref{tab:sim2real}), so all real-robot success comes from
simulated demonstrations alone.}

\subsection{Findings}
\label{sec:findings}

\paragraph{Finding 1: Current policies remain short of reliable deformable manipulation, especially on long-horizon tasks.}
With the same 200 demonstrations per task, $\pi_{0.5}$ achieves the highest average success
rate (77\%), followed by LingBot-VA (70\%), while the remaining policies average 24--50\%
(Table~\ref{tab:main}). The gap is largest on Fold T-shirt, where no policy exceeds 36\%
success; in rollouts, once the folded state deviates or a fold fails, policies rarely
recover.

\paragraph{Finding 2: Appearance shift degrades performance, but some policies already show a degree of robustness to it.}
Changing only the garment's color and texture lowers Flatten T-shirt success by
only 2--14 percentage points for $\pi_{0.5}$, LingBot-VA, and GR00T, but by 28--50 points
for ACT, DP, and Fast-WAM (Table~\ref{tab:shift}). Fast-WAM likely
overfits to the training appearance: it has the highest canonical T-shirt success (92\%) but
the largest drop, and on Flatten silk it falls from 64\% to 0\%.

\paragraph{Finding 3: Policies degrade systematically as the initial deformation increases, with drops far exceeding those under appearance shift.}
The configuration ladder holds the asset, its appearance, and the placement distribution
fixed and varies only the initial deformation of the cloth. On Flatten T-shirt, success
decreases monotonically from C0 to C2 for every policy and reaches 6--14\% at C2, a drop of
58--78 percentage points, compared with 2--50 points under appearance shift; on Flatten
silk, no policy exceeds 2\% at C2 (Table~\ref{tab:shift}). Among policies with nonzero C0 success, the only exception to the larger
drop is Fast-WAM on Flatten silk, which already falls to 0\% under appearance shift. The
deformation state of the object therefore remains a principal challenge for current
policies.

\section{Limitations and Future Work}
\label{sec:limitations}
\benchname{} covers fabric manipulation, namely garments, silk cloth, and fabric bags; ropes,
elastoplastic materials, and volumetric soft bodies are left to future versions. The
configuration ladder consists of three discrete, script-generated, task-specific levels (C0--C2)
rather than a dense sampling of the configuration space; it therefore characterizes how
performance degrades as the initial deformation increases, not performance over the full space
of deformable configurations\rev{, which denser or continuous levels could cover in future work}.
\rev{Future versions will add more complex tasks and evaluate generalization across garment
geometry and fabric material. Separately, we will evaluate generalist policies in a multi-task
setting, training a single policy on the data of all tasks and evaluating it on each task, to
examine whether policies forget individual skills.}

\section{Conclusion}
We presented \benchname{}, a standardized benchmark for bimanual fabric
manipulation with contact-driven grasping, release, and bidirectional
load transfer. Six tasks and controlled configuration and appearance
shifts expose limitations in current policies, particularly on
long-horizon folding and increasingly deformed initial states.
Real-robot evaluation of $\pi_{0.5}$ supports consistency in task
difficulty ordering across simulation and reality. We will release
the simulator, assets, demonstrations, and evaluation code to support
reproducible evaluation.

\appendix
\section{Asset Review and Expansion Details}
\label{app:asset-expansion}

This appendix complements Section~\ref{sec:assets}. Each template asset first passes an
automated review: the mesh loads and renders, its scale and material dependencies are
complete, its deformable simulation remains stable under a standard dynamic check, and its
UV layout agrees with the physical rest panel. The last check is needed because a render
atlas may pack UV charts at arbitrary image-space scales, whereas the rest panel must
preserve the metric lengths and areas of the material.

Expansion varies three factors independently (Table~\ref{tab:asset-expansion-axes}).
\emph{Appearance} variants apply a deterministic recipe of procedural patterns (e.g.,
stripes, checks, dots) and screened printed motifs; a single coverage mask drives
consistent base-color, roughness, normal, and height maps, which affect rendering only.
\emph{Morphology} variants deform semantic regions such as the body, sleeves, collar, and
hem with smooth weights anchored at the seams; the rest pattern is then reconstructed from
the deformed geometry under seam-length constraints rather than by rescaling the UVs, and
the textures are reprojected to the new layout. \emph{Material} variants assign an explicit
profile of density, thickness, anisotropic stretch, shear, and bending stiffness, damping,
and friction. Every build records its recipe, random seed, and source dependencies, so each
released asset can be regenerated exactly, and each passes the same stability check before
release.

\begin{table}[h]
\caption{Independent asset-expansion factors.}
\label{tab:asset-expansion-axes}
\centering
\small
\begin{tabular}{@{}lll@{}}
\toprule
Factor & Changed & Held fixed \\
\midrule
Appearance & render materials and printed patterns & geometry, rest pattern, material \\
Morphology & local geometry and reconstructed rest pattern & appearance recipe, material \\
Material & physical-parameter profile & geometry, rest pattern, appearance \\
\bottomrule
\end{tabular}
\end{table}

\begin{table}[h]
\caption{Template deformable assets used by the six tasks. Extents are the axis-aligned
dimensions of the template mesh before task placement.}
\label{tab:asset-inventory}
\centering
\small
\begin{tabular}{@{}lll@{}}
\toprule
Asset & Tasks & Extents (m) \\
\midrule
Silk & Lift silk, Flatten silk & $0.350\times0.350$ (flat) \\
T-shirt & Flatten T-shirt, Fold T-shirt & $0.744\times0.683\times0.045$ \\
Nonwoven bag & Carry bag, Pack snack & $0.222\times0.383\times0.302$ \\
\bottomrule
\end{tabular}
\end{table}

Pack snack additionally uses a rigid snack bar with a box collision shape, a mass of
0.05\,kg, and a friction coefficient of 0.8.

\section{Simulation Details}
\label{app:sim}

\subsection{Cloth Model and Parameters}
\label{app:sim-cloth}
The in-plane energy per triangle is the anisotropic membrane of
\citet{baraff1998large}: with deformation gradient columns $\mathbf{w}_u,\mathbf{w}_v$
along the warp and weft directions of the flat panel, the stretch conditions are
$C_u = A(\|\mathbf{w}_u\| - 1)$, $C_v = A(\|\mathbf{w}_v\| - 1)$ and the shear
condition $C_s = A\,\mathbf{w}_u^{\top}\mathbf{w}_v$, each with its own stiffness
$k_u, k_v, k_s$ and quadratic energy $\tfrac{1}{2}k C^2$. Bending is the discrete
quadratic energy of \citet{bergou2006quadratic} on each interior edge with an
anisotropic edge stiffness. The three stiffnesses, the bending stiffness, and the
areal density are per-asset material parameters stored with the mesh; the fabric
thickness $t_0$ used by the contact formulation is stored alongside them.

\subsection{Solver}
\label{app:sim-solver}
Each substep solves $(\mathbf{M}/\Delta t^{2} + \mathbf{H})\,\delta\mathbf{x} =
\mathbf{r}$ with a constant $\mathbf{H}$ assembled from the rest-shape elastic
Hessians once at load time, following projective
dynamics~\citep{bouaziz2014projective}. Contact terms (Appendix~\ref{app:sim-contact})
are added to the diagonal of the system at every iteration. The linear system is
solved by conjugate gradients with a Jacobi preconditioner; the number of linear
iterations ramps with the nonlinear iteration index; velocities are updated from the
converged positions. Self-contact uses a BVH broad phase refitted
every substep, vertex--triangle and edge--edge repulsion within a thickness band,
and an edge--face untangling force~\citep{volino2006} that acts only on
already-intersecting pairs.

\subsection{Contact Implementation}
\label{app:sim-contact}
\paragraph{Distance queries.} Each gripper finger shape is voxelised once at load time into a signed distance field with a
per-voxel nearest-triangle table, so the closest point on the shape to any query
point is obtained exactly from the stored triangle without a tree traversal. For a
cloth triangle the minimisation $\min_{\mathbf{x}\in\triangle}\phi(\mathbf{x})$ is
seeded at its three vertices, its centroid, and at up to $n_e$ intersection points
of the rigid mesh's edges with the triangle plane; each seed is refined by a few
projected descent steps inside the triangle and the best point is kept. The edge
seeds are needed because a finger edge of thickness comparable to the mesh edge
length can pass through the interior of a triangle while all four default seeds sit
farther than $h$ from the body.
\paragraph{Normal law.} $f_n = k\,(h-d)$ with $k = E_t A/t_0$. We use $E_t =
20$~kPa, the low-strain transverse compression modulus reported for woven and
knitted fabrics; $t_0$ is read from the asset ($1$~mm for the T-shirt, $2$~mm for
the silk). The force acts at
the closest point along the field gradient and is distributed to the three vertices
with the barycentric weights $w_i$ of that point; the stiffness is added to the
Hessian diagonal with the same weights.
\paragraph{Friction coefficient.} A single cloth-side coefficient is combined
geometrically with the shape coefficient of each rigid surface, and the table and free
objects use the same cloth-side coefficient as the fingers.
\paragraph{Anchored friction.} The anchor is stored in rigid-local coordinates and
transformed to world coordinates as $\mathbf{p}$; $\mathbf{x}_c$ tracks the cloth
material point with barycentric weights fixed at contact formation. The tangential offset
$\mathbf{s} = (\mathbf{I} - \mathbf{n}\mathbf{n}^{\top})(\mathbf{x}_c - \mathbf{p})$
gives the trial force $\mathbf{f}_t = -k_t\mathbf{s}$ with $k_t = \kappa k$, $\kappa = 3$. If
$\|\mathbf{f}_t\| > \mu f_n$, the force is scaled onto the cone and the anchor is
moved so that the stored offset equals the cone limit (return mapping). The anchor is
released when the contact deactivates. The pre-sliding displacement is
$\mu (h-d)/\kappa$, a fraction of the normal compression.
\paragraph{Consistent tangential Hessian.} The anchor stiffness is added to the
Hessian diagonal as the row sum of the full $3\times3$ block, $\sum_j k_t w_i w_j =
k_t w_i$, rather than $k_t w_i^2$, which would under-estimate the stiffness of the
rigid translation mode threefold and make the contact oscillate at the cone limit.
With this lumping the tangential force on a cloth strip resting on a static plate
falls from $7.7$~N to below $0.1$~N.
\paragraph{Deep-penetration guard.} When a triangle's closest point lies deeper than
$0.5$~mm beyond the shell boundary, its vertices are projected back to the shell
after the position update; forces and reaction wrenches are evaluated on the
pre-projection positions so that the projection does not starve the force law.
This is a safeguard for transients, not the mechanism that resolves contact.

\subsection{Co-Simulation}
\label{app:sim-cosim}
After the cloth solve of substep $n$ converges, a kernel re-evaluates every active
contact's normal and tangential force at the converged positions and accumulates
$-\mathbf{f}$ and $-(\mathbf{x}_c - \mathbf{c}_b)\times\mathbf{f}$ per rigid body
$b$ with centre of mass $\mathbf{c}_b$. The wrench is written to the body force
buffer consumed by the MuJoCo solver as an applied external wrench in substep $n+1$.
Wrenches act on the gripper finger links and on free rigid objects; the arm links
are driven by their own actuators. The gripper joints are position actuators with
$k_p = 2\times10^{4}$~N/m, $k_d = 100$~N\,s/m, and a force limit of $6$~N per
finger, so a closed grasp exerts $6$~N regardless of the residual opening. The
stability bound of Sec.~\ref{sec:cosim} evaluates, for the finger of our gripper
($M = 0.030$~kg including armature, $c = 105$~N\,s/m including the velocity gain,
$k_p = 2\times10^{4}$~N/m, $\Delta t = 2$~ms), to $k_{\mathrm{crit}} =
1.15\times10^{5}$~N/m; the closed form, a numerical reproduction of the integrator
update, and the spectral radius of the true one-step map of the rigid-body solver
agree to $0.002\%$. The summed contact stiffness of the linear law during a grasp is
$2$--$4\times10^{4}$~N/m, below this bound.

\subsection{GPU Execution}
\label{app:sim-gpu}
A substep consists of about five hundred small kernels (cloth iterations, collision
queries, rigid-body step, force exchange). In a direct implementation two costs dominate:
host-side kernel-launch latency and the closest-point queries between cloth triangles and
rigid meshes. Three measures address them. (i) The entire substep loop is captured once as
a CUDA graph and replayed, which removes the launch latency; two graphs are kept for odd
substep counts so that the double-buffered state is read consistently, and kinematic
targets are interpolated on the device so that no host read-back breaks the graph.
(ii) Closest-point queries use the per-shape nearest-triangle table of
Appendix~\ref{app:sim-contact} instead of a tree traversal. (iii) The conjugate-gradient
inner loop is fused into a few kernels. Table~\ref{tab:gpuabl} disables each measure in
turn on T-shirt flattening with the full contact stack ($10$ nonlinear iterations, $2$~ms
substep, one H20 GPU with exclusive access). Removing the CUDA graph or the lookup table
each roughly triples the substep time, whereas fusing the conjugate gradients saves
$15\%$. With all three measures in place, the contact formulation takes about $16\%$ of
the GPU time of a substep (measured at $20$ nonlinear iterations); most of the remainder is
the cloth solver's linear solves and the self-collision broad phase.

\begin{table}[h]
\caption{GPU execution ablation on T-shirt flattening with the full contact stack
(9{,}371 vertices, one H20, exclusive access, median of three runs, per-frame
synchronised timing).}
\label{tab:gpuabl}
\centering
\small
\begin{tabular}{@{}lccc@{}}
\toprule
Configuration & ms / substep & Relative & Kernel launches / substep \\
\midrule
Default (graph, fused CG, lookup table) & 6.8  & 1.00 & 0 \\
Without CUDA graph                       & 19.8 & 2.92 & 519 \\
Without fused conjugate gradients         & 7.8  & 1.15 & 0 \\
Without nearest-triangle lookup table     & 20.4 & 3.00 & 0 \\
\bottomrule
\end{tabular}
\end{table}

\subsection{Ablation Details}
\label{app:sim-ablation}
\paragraph{Criteria of Table~\ref{tab:ablation}a.} Each cell uses 20 seeds of scripted
trajectories. A finger--fabric \emph{intersection event} is a run of at least 10
consecutive frames, with the jaws at least 90\% open, in which a triangle of the
finger mesh intersects a cloth triangle. \emph{Release} (T-shirt): after the jaws
open, the fabric falls back as defined in Appendix~\ref{app:sim-criteria}, so fabric
carried up on a finger counts as a failure whether or not it intersects the finger. \emph{Penetration} (silk): an intersection event occurs after
the jaws open. \emph{Lift} (pack): when the bag is raised by more than 5~cm, the packed item
rises by at least half the bag's rise. Runs that diverge or fail inverse kinematics
count as failures.

\paragraph{Edge seeds.} Removing the rigid-edge seeds of the closest-point search
(20 seeds per cell) leaves grasp, hold, and the Pack snack task unchanged; on the
T-shirt an intersection event follows opening in 2 of 20 runs, and a finger that
pierces the fabric on opening stays inside it until separation in 3 of 20, against
none with the full model.

\begin{figure}[htbp]
\centering
\includegraphics[width=\linewidth]{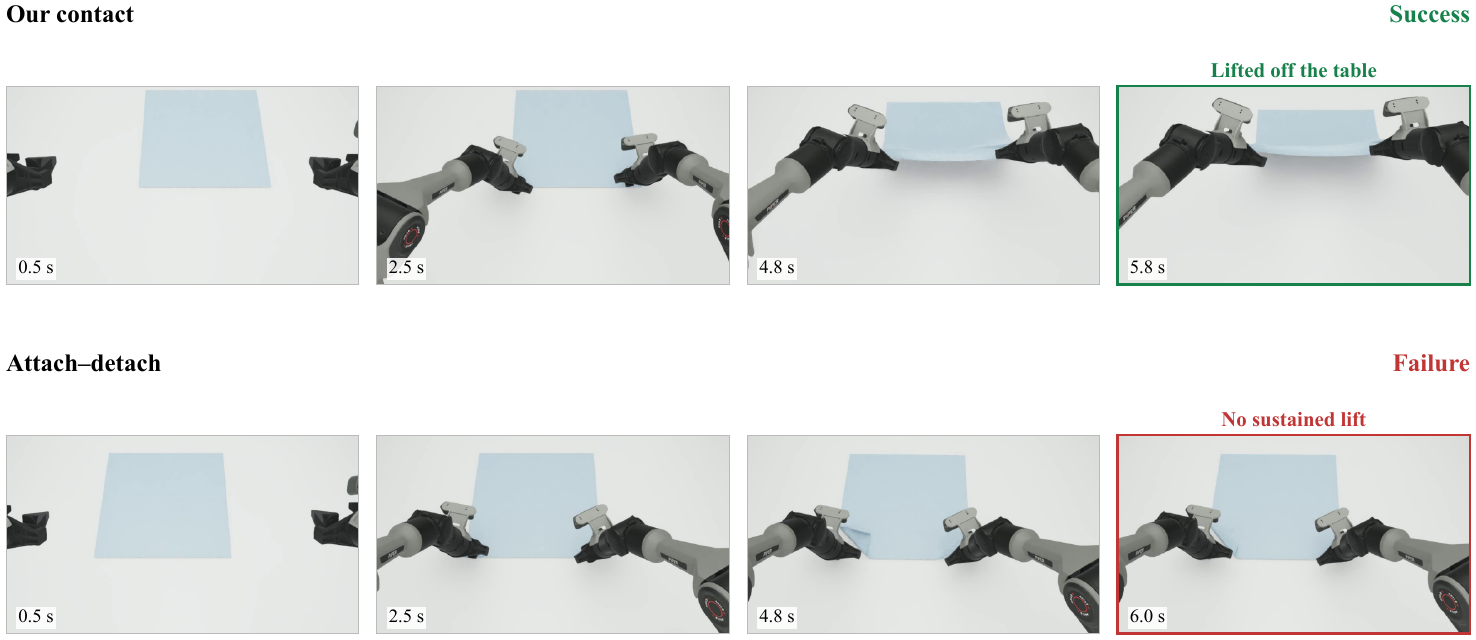}
\caption{\textbf{Qualitative grasp outcomes under learned-policy control.}
Top: our contact maintains the grasp and lifts the silk off the table
(\textcolor{green!45!black}{success}). Bottom: attach--detach fails to maintain
an effective grasp (\textcolor{red!70!black}{failure}). Frames progress from
approach to the final outcome; timestamps are relative to the beginning of each
source clip, and the two rows are not synchronized replays. These examples
illustrate the grasp behavior; success rates are reported in
Table~\ref{tab:ablation}b.}
\label{fig:policy-contact-keyframes}
\end{figure}

\paragraph{Closed-loop policy examples.}
\label{app:policy-contact}
Figure~\ref{fig:policy-contact-keyframes} complements the aggregate comparison in
Table~\ref{tab:ablation}b with representative Lift silk sequences. With our contact,
the grippers lift the fabric clear of the table. In the attach--detach example,
the grasp is disrupted and no sustained lift is achieved. The images show the
visible grasp outcomes; the threshold-triggered detachment mechanism is discussed
in Section~\ref{sec:validation}.

\subsection{Behavioural Criteria}
\label{app:sim-criteria}
All grasp-related criteria are binary per hand and computed from cloth vertex
positions and gripper state only, so they apply to any simulator.
\emph{First grasp}: after the first close command the grasp point rises with the
hand by at least half the commanded lift, without retry.
\emph{Hold}: the grasp point stays within $30$~mm of the tool centre point from the
first grasp until the open command, throughout carrying or flinging.
\emph{Release}: once the jaws open, the fabric falls back instead of being carried by
the fingers, i.e., the highest cloth vertex stays within $5$~mm of its height at the moment
of opening. Table~\ref{tab:xsim}, whose trials end with the arms retracted, additionally
requires that no cloth vertex lies within $3$~cm of either jaw centre at the end of the
trial. Both tests use cloth vertices only and therefore apply to every simulator; the same
definition is used in Table~\ref{tab:ablation}a.
\emph{Success}: grasp, hold, and release all pass and the trial does not diverge.
\emph{Diverged}: the cloth velocity exceeds a fixed threshold or a non-finite state appears.
Counts are reported per trajectory and never averaged, because release failures are
intermittent rather than noisy.

\paragraph{Cross-simulator details.} Table~\ref{tab:xsim-detail} breaks down
Table~\ref{tab:xsim}. The step times in Table~\ref{tab:xsim} are measured on one RTX~4090
with the 9{,}371-vertex T-shirt as wall-clock time per physical step, each simulator in
the configuration of Table~\ref{tab:xsim} (Newton VBD at twice its released substep rate,
see below), as the median of three runs excluding rendering and initialization;
they are not comparable with the H20 timings of Table~\ref{tab:gpuabl}.
\emph{Genesis.} Under the same protocol, with both Piper arms coupled to the cloth,
Genesis~\citep{genesis2024} (v1.4.2, IPC cloth, parameters of its robot--cloth example)
takes 35.9~ms per step, $7.5\times$ our contact (4.8~ms). The grasp does not form in
Genesis, so this is a lower bound.
\emph{Isaac Sim.} The fingers pass through the fabric while closing, so it is not held
between the jaws; one position forms a single-hand grasp. In a separate sweep,
commanding the fingertips 15~mm below the table, which is unsafe on the real robot,
forms a grasp at only 3 of 19 positions, and the outcome flips under 2~cm shifts.
\emph{Newton VBD.} At the released 1440~Hz step the simulation diverges in 39 of 45
trials, during closing or the fling. Doubling the iteration count does not remove it;
halving the step reduces it to 12 of 45 (the setting in Table~\ref{tab:xsim}). Trials that
hold never release: the fingers remain inside the fabric when the jaws open.
\emph{Newton Style3D.} The released contact holds in every trial, but the fingers remain
inside the fabric and it never releases; it releases only if collisions are disabled at
opening. \rev{Figure~\ref{fig:xsim-real} shows these grasp and release failures next to the
real robot.}

\rev{\paragraph{Provenance of the test trajectory.} The replayed trajectory is not recorded in
\benchname{}. It comes from a generic, simulator-agnostic grasp planner: a grasp point is fixed on
the cloth, approach, jaw closing, lift, hold, and jaw opening are specified as end-effector
waypoints, and the joint commands are solved by inverse kinematics. The gripper receives the same
full-closing command, with the same width and speed, in every simulator. The cloth placement is
shifted over the fixed $5\times3$ grid with 2~cm steps, each placement is replayed three times,
and all 45 trials are reported without exclusion. Neither the planner, the waypoints, nor the
gripper commands were adjusted based on any simulator's outcome; the only adjustment favors a
baseline (the Isaac Sim fingertip offset above). The same trajectory grasps, holds, and releases
the T-shirt on the real robot (Appendix~\ref{app:xsim-real}), so the baseline failures are not
caused by an infeasible trajectory. Three additional grasp trajectories teleoperated by a human
in simulation give the same outcome in every simulator: the fingers penetrate the cloth in
Newton VBD and Style3D, and Isaac Sim PBD rarely forms a grasp.}

\subsection{\rev{Sim--Real Replay of the Contact Trajectory}}
\label{app:xsim-real}
\rev{The trajectory of Table~\ref{tab:xsim} is executed on the real dual-arm robot and replayed in
each simulator with the same T-shirt in the same initial pose (Figure~\ref{fig:xsim-real}); the
full side-by-side videos are included in the supplementary material.}

\newcommand{\xsrimg}[2][white]{{\setlength{\fboxsep}{0pt}\setlength{\fboxrule}{1pt}\fcolorbox{#1}{white}{#2}}}
\begin{figure}[!t]
\centering
\revon
\setlength{\tabcolsep}{0.5pt}
\renewcommand{\arraystretch}{0.5}
\begin{tabular}{@{}r ccccc@{}}
 & \scriptsize Real robot & \scriptsize Ours & \scriptsize Isaac Sim PBD & \scriptsize Newton VBD & \scriptsize Newton Style3D \\
\rotatebox{90}{\scriptsize\hspace{14pt}Grasp} & \xsrimg{\includegraphics[width=0.185\linewidth]{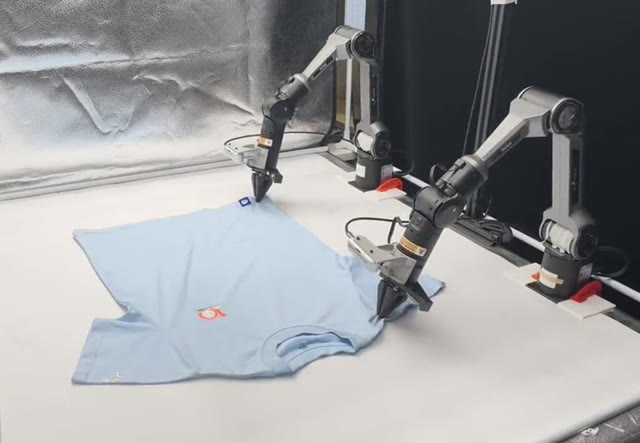}} & \xsrimg{\includegraphics[width=0.185\linewidth]{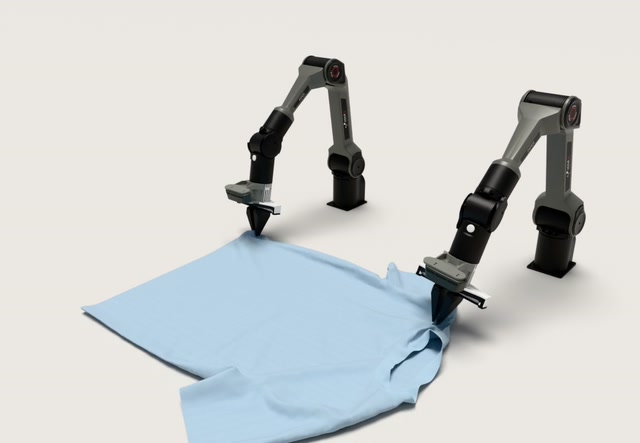}} & \xsrimg{\includegraphics[width=0.185\linewidth]{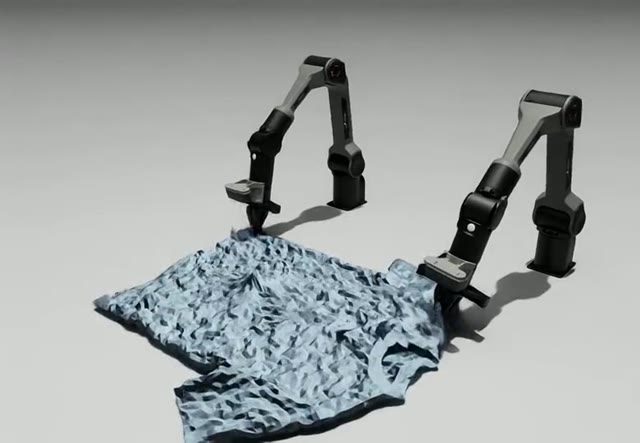}} & \xsrimg{\includegraphics[width=0.185\linewidth]{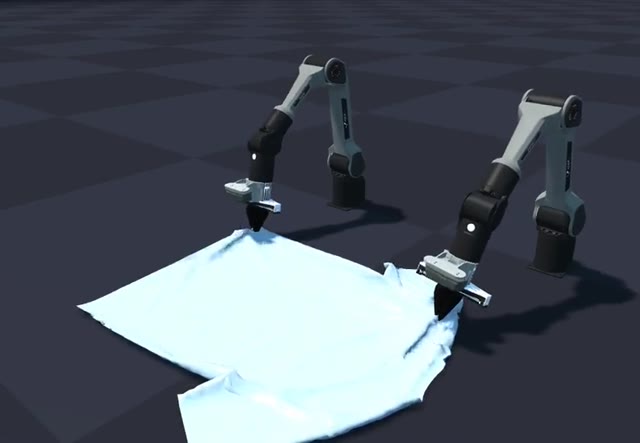}} & \xsrimg{\includegraphics[width=0.185\linewidth]{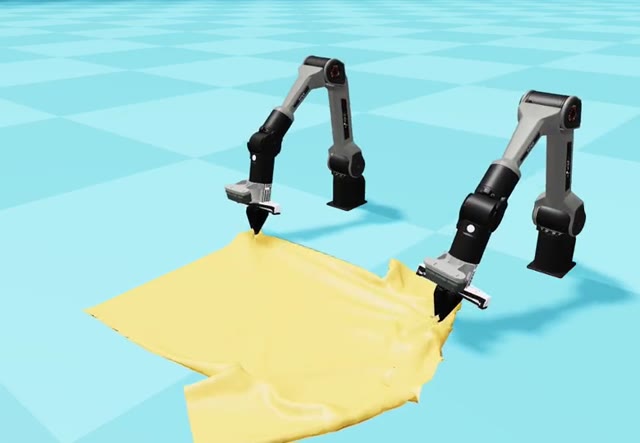}} \\
\rotatebox{90}{\scriptsize\hspace{8pt}Lift, hold} & \xsrimg{\includegraphics[width=0.185\linewidth]{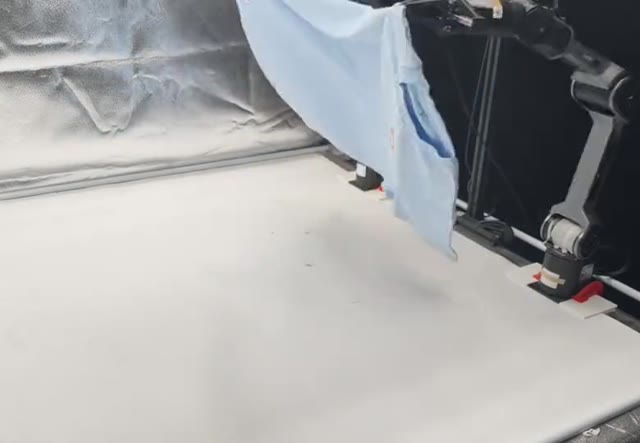}} & \xsrimg{\includegraphics[width=0.185\linewidth]{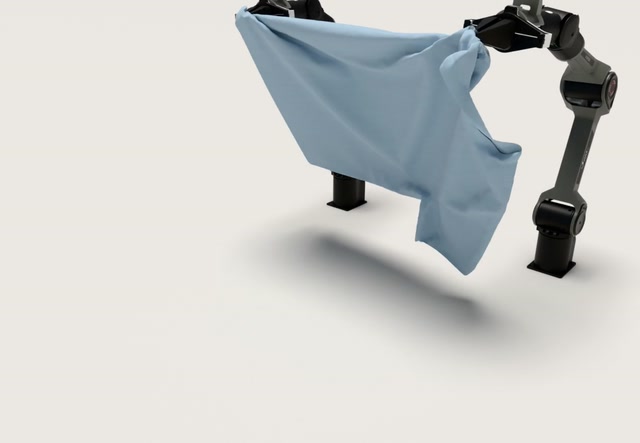}} & \xsrimg[red]{\includegraphics[width=0.185\linewidth]{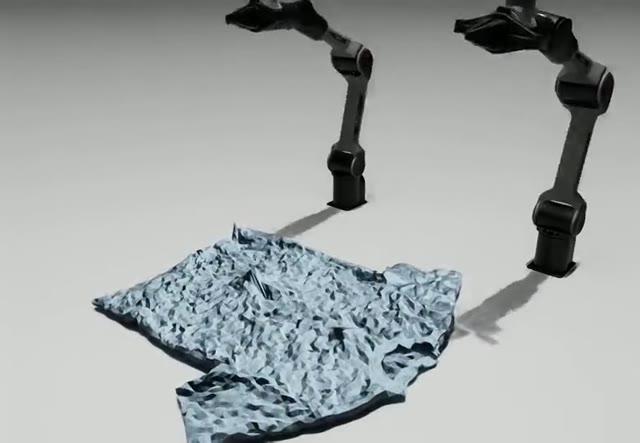}} & \xsrimg{\includegraphics[width=0.185\linewidth]{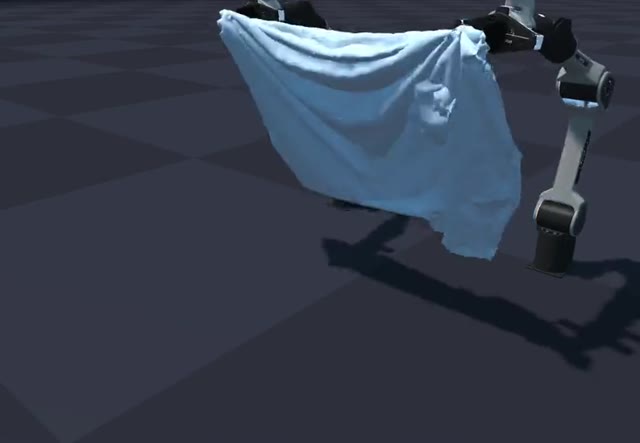}} & \xsrimg{\includegraphics[width=0.185\linewidth]{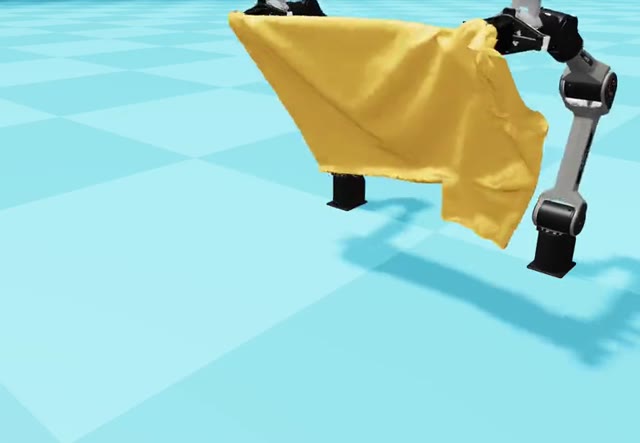}} \\
\rotatebox{90}{\scriptsize\hspace{1pt}After release} & \xsrimg{\includegraphics[width=0.185\linewidth]{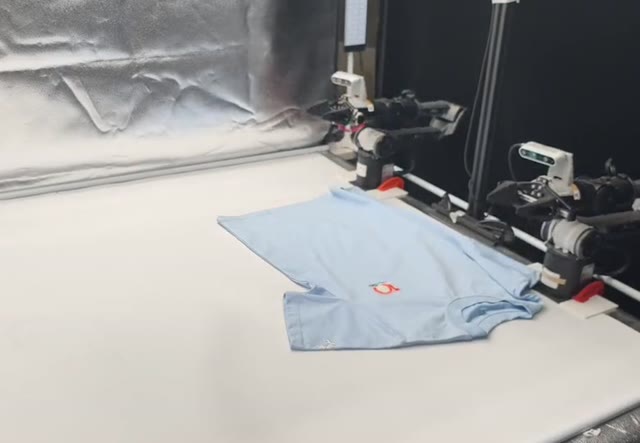}} & \xsrimg{\includegraphics[width=0.185\linewidth]{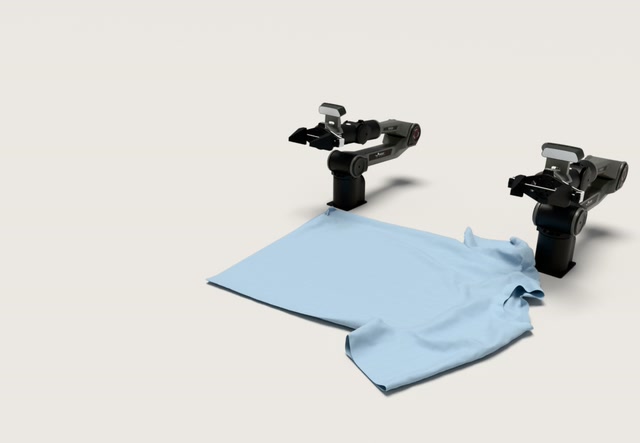}} & \xsrimg{\includegraphics[width=0.185\linewidth]{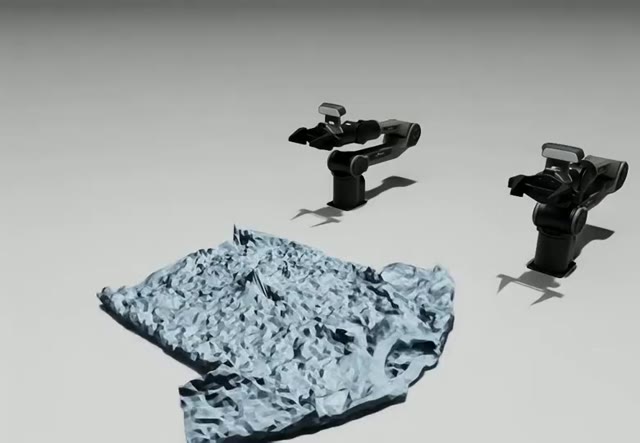}} & \xsrimg[red]{\includegraphics[width=0.185\linewidth]{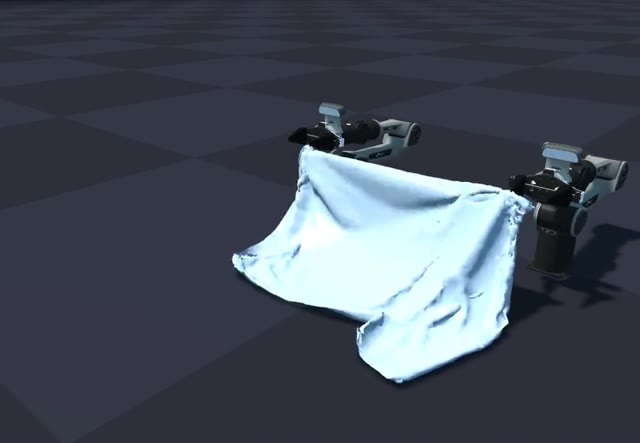}} & \xsrimg[red]{\includegraphics[width=0.185\linewidth]{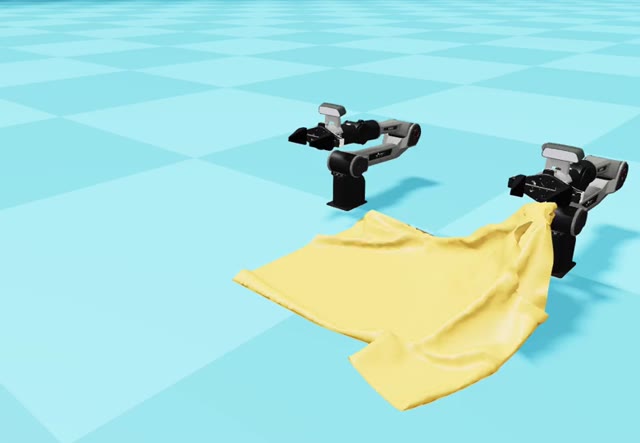}} \\
\end{tabular}
\caption{\textbf{The same grasp--hold--release trajectory on the real robot and in four
simulators}, at $t=3$, 6, and 10~s on a shared timeline; frames are taken from the
side-by-side videos in the supplementary material. Only our contact matches the real robot in
all three stages. Red: Isaac Sim PBD forms no grasp and the T-shirt stays on the table; in Newton VBD and Style3D the fingers
penetrate the T-shirt, which remains on them after the jaws open.}
\label{fig:xsim-real}
\end{figure}

\begin{table}[!htbp]
\small
\caption{Per-simulator breakdown (45 trials each; same replay as Table~\ref{tab:xsim}).}
\label{tab:xsim-detail}
\setlength{\tabcolsep}{3pt}
\begin{tabular*}{\linewidth}{@{\extracolsep{\fill}}l cccc@{}}
\toprule
 & Grasp & Hold & Release & Diverged \\
\midrule
Isaac Sim, PBD cloth                   & 3/45  & 0/45  & --    & 0/45  \\
Newton VBD, 1440~Hz                    & 30/45 & 27/45 & 0/45  & 39/45 \\
Newton VBD, 1440~Hz, 24 / 48 iter.\textsuperscript{a} & -- & -- & -- & 3/3, 3/3 \\
Newton VBD, 2880~Hz                    & 39/45 & 39/45 & 0/45  & 12/45 \\
Newton VBD, 5760~Hz\textsuperscript{a} & --    & --    & --    & 0/3   \\
Newton Style3D                         & 45/45 & 45/45 & 0/45  & 0/45  \\
Newton Style3D, our contact            & 45/45 & 45/45 & 45/45 & 0/45  \\
\bottomrule
\end{tabular*}
\par\smallskip\footnotesize\textsuperscript{a}Three positions only.
\end{table}

\needspace{6\baselineskip}
\subsection{\rev{Sensitivity of Task Outcomes to Material Parameters}}
\label{app:sim-material}
\rev{Each asset carries its own material parameters (Appendix~\ref{app:sim-cloth}), so we check how
sensitive the benchmark tasks are to them. We scale the stretch stiffness, bending stiffness, and
areal density of the fabric over $0.5$--$2\times$. Task outcomes of grasp-and-place and folding are
largely insensitive to these parameters, as grasping, holding, and releasing are governed by the
contact mechanism within the tested ranges. Dynamic flinging in the flattening tasks is sensitive
mainly to air damping rather than to these parameters. In Carry bag, the handle stiffness is
additionally randomized in both training and evaluation (Appendix~\ref{app:tasks}).}

\section{Configuration Ladder: Generation Details}
\label{app:ladder-stats}

Each level is generated inside the task environment by a deterministic program whose
random choices are driven by the episode seed. For silk, C0 selects one of the two far
corners and folds it inward; its sampled displacement is 0.15--0.20\,m in the long
direction and 0.05--0.15\,m laterally. C1 samples a fold count uniformly from one or
two, samples that many distinct corners uniformly without replacement from all four
corners, and applies the same inward fold sequentially in the sampled order. C2 samples
two mesh pins with a separation of 0.10--0.45\,m, raises the cloth centroid to
0.32--0.48\,m with a 0.15--0.25\,m horizontal offset, holds for 1--3\,s, releases
the cloth, settles it for 2\,s, recenters it, and settles it for another 0.3\,s. The
silk C0/C1 fold primitive executes three segments of 0.4, 0.6, and 0.4\,s, places the
selected vertex at 0.005\,m, pins it for 0.3\,s, then releases and settles for 1.5\,s.

For the T-shirt, C0 is a right-sleeve-corner inward fold, C1 is a right-half inward
fold, and C2 is a randomly selected deep fold over the four corner regions.
Figure~\ref{fig:ladder} shows representative states. Within each ladder, the asset
identity, physics, task parameters, and planar placement distribution are held fixed.

\section{Sim-to-Real Cells}
\label{app:sim2real}
\begin{table}[h]
\caption{Per-cell numbers behind Figure~\ref{fig:sim2real}: sim-to-real
difficulty ordering with $\pi_{0.5}$. Simulation cells use $N{=}50$ and
completed real-robot cells use $N{=}20$. Simulation rates match
Table~\ref{tab:main}. \rev{Pretrained: $\pi_{0.5}$ without fine-tuning on \benchname{},
evaluated on the real robot under the same success criteria.}}
\label{tab:sim2real}
\centering
\small
\begin{tabular}{@{}lccc>{\revon}c@{}}
\toprule
Task & Sim SR (\%) & Real SR (\%) & Real trials & Real SR, pretrained (\%) \\
\midrule
Lift silk & 98 & 70 & 20 & 0 \\
Flatten T-shirt & 82 & 55 & 20 & 0 \\
Flatten silk & 86 & 65 & 20 & 0 \\
Carry bag & 66 & 50 & 20 & 0 \\
Pack snack & 92 & 85 & 20 & 0 \\
Fold T-shirt & 36 & 35 & 20 & 0 \\
\midrule
\multicolumn{5}{@{}l}{Spearman $\rho=0.94$; bootstrap 95\% CI: $[0.26,\,0.99]$} \\
\multicolumn{5}{@{}l}{Kendall $\tau_b=0.87$; bootstrap 95\% CI: $[0.20,\,0.97]$} \\
\bottomrule
\end{tabular}
\end{table}

The correlation intervals use 20,000 within-cell episode bootstrap replicates
(random seed 20260924), resampling binary outcomes independently in simulation
and on the real robot while holding the six tasks fixed. We report percentile
95\% intervals; these quantify episode-sampling uncertainty for the evaluated
tasks rather than uncertainty over a larger population of tasks.

\section{Task Definitions and Success Criteria}
\label{app:tasks}

\subsection{Coverage of Physical Regimes}

The five regimes in Section~\ref{sec:task-design} are overlapping physical requirements rather than a
partition of the task set. Table~\ref{tab:task-regime-map} makes the intended mapping
explicit. A mark denotes a primary capability exercised by the task, not a claim that
all other physical effects are absent.

\begin{table}[h]
\caption{Mapping from the six tasks to the five physical regimes. SC: self-contact;
CR: recovery under self-occlusion; MT: dynamic momentum transfer; FH:
friction-limited hold without attachment; VB: volumetric-bag load transmission.}
\label{tab:task-regime-map}
\centering
\small
\begin{tabular}{@{}lccccc@{}}
\toprule
Task & SC & CR & MT & FH & VB \\
\midrule
Fold T-shirt    & \cmark &          &          &          &          \\
Flatten T-shirt &        & \cmark   & \cmark   &          &          \\
Flatten silk    &        & \cmark   & \cmark   &          &          \\
Lift silk      &        &          &           & \cmark   &          \\
Carry bag       &        &          &           & \cmark   & \cmark   \\
Pack snack            &        &          &           & \cmark   & \cmark   \\
\bottomrule
\end{tabular}
\end{table}

\subsection{Evaluator Semantics}

The evaluator observes the simulation at a fixed polling interval (1\,s). Unless a task states otherwise, a \emph{sustained}
predicate means that it is true at two consecutive evaluation polls. ``Ever'' means true at any poll;
``end-held'' means true at the final two polls. Threshold comparisons are strict when
the implementation uses $<$ or $>$ and inclusive when it uses $\leq$ or $\geq$.
Continuous scores are independent of the binary success flag and remain defined for
failed episodes.

\paragraph{Lift silk.}
At a poll, one hand holds the silk if the nearest cloth particle is less than
0.05\,m from its end effector and that particle is above 0.10\,m. The whole cloth is
lifted if its mean particle height is above 0.05\,m. Success requires both hand-hold
predicates and the whole-cloth predicate for two consecutive polls. Each hand
contributes 50 points if it is end-held, 25 if it was held at least once but is not
end-held, and 0 otherwise. The two hand scores are summed.

\paragraph{Flatten silk.}
After the policy horizon, both grippers open and the cloth settles for 2\,s. Coverage
is the union area of table-plane particle footprints with radius 0.005\,m, divided by
the corresponding area of the canonical flat mesh and clipped to one for scoring.
Success requires final coverage at least 0.95 and maximum particle height below
0.05\,m. The score is 100 times final settled coverage. The full unclipped coverage
and height series are retained.

\paragraph{Flatten T-shirt.}
The evaluator combines projected coverage $c$ with a planar, rigidly aligned distance
$d_w$ to the canonical flat mesh. The distance is a weighted bidirectional
correspondence measure after iterative rigid alignment; lower is better. The current
success rule is
\[
  c \geq 0.60, \qquad d_w \leq 0.25, \qquad \bar z < 0.03\,\mathrm{m}.
\]
The score is
\[
100\left[0.4\,\operatorname{clip}\!\left(\frac{c-0.25}{0.78-0.25},0,1\right)
+0.6\,\operatorname{clip}\!\left(\frac{0.36-d_w}{0.36-0.196},0,1\right)\right].
\]
The episode ends at the first successful poll. Initial, final, and canonical particle
arrays are stored to permit offline rescoring.

\paragraph{Fold T-shirt.}
The evaluator freezes a garment-local frame at the first poll and awards 30 points for
the left side fold, 30 for the right side fold, and 40 for the forward fold. A side is
latched when its sleeve-center and hem keypoints are both within 0.35 of their initial
lateral distance from the torso centerline and rest below 0.10\,m. The forward stage
is latched when the mean hem keypoint has passed the torso center by at least 0.10 of
the initial garment length and rests below 0.14\,m. The final 40 points and binary
success additionally require a settled envelope: mean height below 0.09\,m, maximum
height below 0.30\,m, footprint width no greater than 0.60 of its initial value, and
footprint length no greater than 0.65 of its initial value. Success requires all three
stages and the final envelope. Both grippers are released before a 2\,s terminal
settle.

\paragraph{Carry bag.}
The handle position, handle stiffness, and bag appearance are randomized in both training and
evaluation. Evaluation uses the ten training-pool appearances, with five episodes per
appearance and held-out initialization seeds. All vertical displacements are measured
relative to the first evaluation poll. A
side is valid when the mean height of its strap tip and mid vertices rises by more
than 0.15\,m and the nearest strap vertex is less than 0.02\,m from the corresponding
TCP pinch center. The bag predicate requires its mean particle height to rise by more
than 0.03\,m. Success requires both side predicates and the bag predicate at two
consecutive polls. Each side contributes 50 points if sustained, 25 if true at only a
nonsustained poll, and 0 otherwise; bag lift gates success but does not add score. The
episode ends at the first sustained lift window: Carry bag success is defined by
friction-limited handle acquisition and sustained lifting, not by arrival at a separate
transport target.

\paragraph{Pack snack.}
Relative to the first poll, the rigid bar is ``up'' after rising by more than
0.10\,m. It is ``in bag'' when its table-plane distance from the dynamically estimated
bag-mouth center is less than 0.12\,m and its height is at least 0.02\,m below the
estimated mouth rim. Each strap is valid after rising by more than 0.10\,m while
remaining within 0.02\,m of its TCP pinch center. Loaded carry requires the bag mean
height to rise by more than 0.05\,m while the bar remains in the bag and rises by at least
half the bag's rise, so a bar left on the table under the lifted bag does not count. The score assigns
20 points to bar lift, 30 to insertion, 20 to each valid strap, and 10 to loaded
carry, with every stage requiring two consecutive polls. Success requires sustained
insertion and a sustained joint window in which both straps and loaded carry are
valid.

\paragraph{Process traces and failure labels.}
Besides the binary outcome and the score, each episode logs process traces (coverage over
time, semantic-keypoint errors, grasp duration, and stage-completion indicators), which
distinguish policies that fail at different points even when their success rates are
identical. Failures are labeled by observed behavior: missed
acquisition, acquisition followed by early slip, incomplete fold or flatten stage,
failure to settle, failed insertion/containment, one-sided handle acquisition, empty
bag lift, and loss during carry. These labels describe observed behavior only and do
not infer an unmeasured physical cause. The raw per-poll stage series is retained so
that labels can be reproduced or refined offline.

\subsection{Demonstration Generation and Data Format}
\label{app:data-interface}

Each task provides a scripted expert expressed through semantic asset keypoints,
object-relative targets, and reusable bimanual motion primitives. At the start of an
episode, the task randomizer samples the placement and initial state from a seeded
distribution, and the planner instantiates the expert trajectory for that scene;
collection and evaluation therefore share the same physics, control frequency,
randomization, and success checker. Collection seeds are disjoint from evaluation seeds.

Demonstrations are stored in LeRobot format at 25\,Hz. Each frame contains three RGB
streams, one top camera and two wrist cameras, rendered at $1280\!\times\!720$. State and
action are both 14-dimensional: six joints and one gripper value per arm, with the gripper
normalized to 0 closed and 1 open.

\paragraph{Cameras.} The three simulated cameras replicate the Intel RealSense D435i
cameras of the real dual-arm robot and are identical across tasks. The top camera is fixed
in the world about $0.77$~m above the table, looking forward and $60^\circ$ below the
horizontal. Each wrist camera is rigidly attached to the last link of its arm at the pose
obtained from the real robot's calibration, with its optical axis $20^\circ$ from the
link axis. All three are ideal pinhole cameras with a $70.3^\circ\times43.2^\circ$ field of
view (focal length $\approx 908$~px at $1280\!\times\!720$, principal point at the image
centre) and are rendered with path tracing. The real cameras stream at the same resolution,
and policies receive images at 25\,Hz in both settings.

\section{Policy Training and Unified Evaluation Protocol}
\label{app:policy-protocol}

The benchmark covers ACT, Diffusion Policy (DP), $\pi_{0.5}$, GR00T, LingBot-VA, and
Fast-WAM. Table~\ref{tab:training-recipes} records the current reference recipes. The
scheduled terminal checkpoint is evaluated; any result from an earlier checkpoint is
marked in the corresponding table.
When the number of devices changes, steps, per-device batch size, or gradient
accumulation are adjusted to preserve the total number of training samples; model
revisions and algorithm-specific augmentation remain fixed.

\begin{table}[h]
\caption{Reference training recipes. ``Effective batch'' includes all devices and
gradient accumulation.}
\label{tab:training-recipes}
\centering
\small
\begin{tabular}{@{}llll@{}}
\toprule
Policy & Schedule & Effective batch / devices & Notes \\
\midrule
ACT & 30k steps & 16 on one device & save every 5k \\
DP & 12.5k one-device equivalent & 64 & 6.25k at two devices \\
$\pi_{0.5}$ & 30k steps & 8 & bf16, grad. checkpointing \\
GR00T & 20k steps & global batch 32 & 16-step prediction horizon \\
LingBot-VA & 30k steps & 8 devices, effective 8 & task-specific action quantiles \\
Fast-WAM & 5 epochs & 8 devices, 16/device & ZeRO-1; no CPU offload \\
\bottomrule
\end{tabular}
\end{table}

Fast-WAM inference uses action horizon
32, replans every 32 actions, and uses 20 denoising steps unless a run manifest
explicitly overrides them. GR00T predicts a 16-step horizon by default.

Each principal task--condition--policy cell targets $N=50$ completed episodes from
one fixed evaluation-seed schedule shared across policies. This is 50 total episodes,
not 50 episodes for each of three seeds. The runner resumes incomplete cells and
retries failed workers; an episode JSON that is absent is not counted in $N$. Success
rates are reported with two-sided Wilson 95\% confidence intervals
(Tables~\ref{tab:main-ci} and~\ref{tab:shift-ci}); the mean score is reported as a sample
mean. With $N{=}50$, an interval spans roughly $\pm14$ points at a 50\% success rate and
narrows toward the extremes.

\paragraph{Success-rate uncertainty.} Tables~\ref{tab:main-ci} and~\ref{tab:shift-ci}
complement the point estimates in the main results with confidence intervals for each
policy--task condition. Table~\ref{tab:main-ci} covers the standard task distributions,
while Table~\ref{tab:shift-ci} covers configuration and appearance shifts. With 50
episodes per cell, these intervals highlight the uncertainty in small success-rate
differences.

\begin{table}[!ht]
\caption{Wilson 95\% confidence intervals (\%) for the success rates of
Table~\ref{tab:main} ($N{=}50$ per cell).}
\label{tab:main-ci}
\centering
\footnotesize
\setlength{\tabcolsep}{2pt}
\begin{tabular*}{\linewidth}{@{\extracolsep{\fill}}l cccccc@{}}
\toprule
Policy & Lift silk & Flatten silk & Flatten T-shirt & Carry bag & Pack snack & Fold T-shirt \\
\midrule
$\pi_{0.5}$ & 90--100 & 74--93 & 69--90 & 52--78 & 81--97 & 24--50 \\
LingBot-VA & 52--78 & 74--93 & 74--93 & 37--63 & 90--100 & 24--50 \\
Fast-WAM & 44--71 & 50--76 & 81--97 & 67--89 & 3--19 & 0--7 \\
GR00T & 56--81 & 33--60 & 58--83 & 28--54 & 40--67 & 0--10 \\
ACT & 17--42 & 4--21 & 56--81 & 26--52 & 39--65 & 19--44 \\
DP & 26--52 & 0--7 & 65--87 & 3--19 & 10--31 & 0--7 \\
\bottomrule
\end{tabular*}
\end{table}

\begin{table}[!ht]
\caption{Wilson 95\% confidence intervals (\%) for the success rates of
Table~\ref{tab:shift} ($N{=}50$ per cell).}
\label{tab:shift-ci}
\centering
\footnotesize
\setlength{\tabcolsep}{1.5pt}
\begin{tabular*}{\linewidth}{@{\extracolsep{\fill}}l cccc cccc@{}}
\toprule
 & \multicolumn{4}{c}{Flatten T-shirt} & \multicolumn{4}{c}{Flatten silk} \\
\cmidrule(lr){2-5}\cmidrule(l){6-9}
Policy & C0 & C1 & C2 & C0 + app. & C0 & C1 & C2 & C0 + app. \\
\midrule
$\pi_{0.5}$ & 69--90 & 31--58 & 6--24 & 63--86 & 74--93 & 11--33 & 0--10 & 50--76 \\
LingBot-VA & 74--93 & 48--74 & 4--21 & 71--92 & 74--93 & 13--35 & 0--7 & 74--93 \\
Fast-WAM & 81--97 & 22--48 & 7--26 & 29--56 & 50--76 & 8--29 & 0--7 & 0--7 \\
GR00T & 58--83 & 22--48 & 7--26 & 44--71 & 33--60 & 10--31 & 0--7 & 26--52 \\
ACT & 56--81 & 8--29 & 2--16 & 17--42 & 4--21 & 2--16 & 0--7 & 1--13 \\
DP & 65--87 & 21--46 & 4--21 & 37--63 & 0--7 & 0--7 & 0--7 & 0--7 \\
\bottomrule
\end{tabular*}
\end{table}

\section{\rev{Comparison with Existing Benchmarks}}
\label{app:bench-compare}
\revon
Table~\ref{tab:bench-compare} compares \benchname{} with existing simulated benchmarks for
deformable manipulation (Section~\ref{sec:rw-bench}). These suites complete their tasks, but
through attachment, adhesion, pick-and-place primitives, or task-specific grasp strategies. Under
the same parallel-jaw grasp, Isaac Sim PBD, on which the most recent suites are built, forms no
grasp, and the simulators that do hold the cloth fail to release it (Table~\ref{tab:xsim}), so
outcomes partly reflect the grasp model rather than the policy (Table~\ref{tab:ablation}b). In
\benchname{}, grasp, release, and the transfer of a bag's load arise from contact forces alone,
and policies are evaluated along deformation-state and appearance axes separately, with a
stage-wise progress score in addition to the success rate. Rigid-object suites such as
RoboTwin~2.0~\citep{chen2025robotwin2}, LIBERO~\citep{liu2023libero}, and
RoboDojo~\citep{robodojo2026} are not listed, as they contain no deformable task or a single one.

\begin{table}[H]
\revon
\caption{Comparison with simulated benchmarks for deformable manipulation. Contact-driven grasp
and release: whether grasping and releasing the fabric arise from contact forces between the
gripper and the fabric. Reported generalization axes: held-out axes evaluated in the respective
paper.}
\label{tab:bench-compare}
\centering
\footnotesize
\setlength{\tabcolsep}{2.5pt}
\begin{tabular}{@{}l l l >{\raggedright\arraybackslash}p{2.4cm} l@{}}
\toprule
Benchmark & \makecell[l]{Contact-driven\\grasp \& release} & \makecell[l]{Bag with\\contents} & \makecell[l]{Reported\\generalization axes} & Policies \\
\midrule
SoftGym~\citep{lin2020softgym}                 & \xmark\ attachment & \xmark & -- & RL \\
Deformable Ravens~\citep{seita2021deformable}  & \xmark\ attachment & \makecell[l]{partial\\(attached bag)} & -- & IL \\
DaXBench~\citep{chen2023daxbench}              & \xmark\ primitives & \xmark & -- & \makecell[l]{RL, IL,\\planning} \\
GarmentLab~\citep{lu2024garmentlab}            & \xmark\ attachment & \xmark & -- & RL \\
DexGarmentLab~\citep{wang2025dexgarmentlab}    & \xmark\ PBD\textsuperscript{a} & \xmark & unseen garments & IL \\
LeHome~\citep{lehome2026}                      & \xmark\ PBD\textsuperscript{a} & \xmark & -- & IL, VLA \\
FlatLab~\citep{flatlab2026}                    & \xmark\ PBD\textsuperscript{a} & \xmark & unseen objects / categories & IL, VLA \\
\midrule
\benchname{}                                   & \cmark & \cmark & deformation state; appearance & IL, VLA, WAM \\
\bottomrule
\end{tabular}
\par\smallskip\footnotesize\textsuperscript{a}Isaac Sim PBD cloth, which forms no grasp in our
cross-simulator replay (Table~\ref{tab:xsim}).
\end{table}

\section{\rev{Quantitative Sim-to-Real Comparison on RGBench}}
\label{app:rgbench}
\revon
RGBench~\citep{hu2026rgbench} is a public sim-to-real benchmark for garment manipulation. In its
robot mode, the recorded joint positions and gripper openings of real bimanual fold and
grasp-and-lift trajectories are replayed in simulation from the initial garment state of each real trial,
and the per-frame Chamfer distance ($L_1$, real to simulated point cloud) is averaged over the
trajectory; lower is better. Grasp-and-lift additionally tests whether the grip holds the garment
without slipping, as on the real robot. We run our simulator with the contact parameters of the
main evaluation (Section~\ref{sec:validation}) and without per-garment tuning, with the arm
and gripper PD-controlled and force-limited as in our tasks. Baseline numbers are those reported
by RGBench for the publicly available simulators, over the seven garments on which they are
reported. Table~\ref{tab:rgbench} lists the per-garment results.

\begin{table}[H]
\revon
\caption{Per-garment Chamfer distance on RGBench (robot mode; lower is better). Best per row in
bold. Baseline numbers are as reported by RGBench.}
\label{tab:rgbench}
\centering
\small
\begin{tabular}{@{}ll ccc@{}}
\toprule
Garment & Motion & PyBullet & Isaac Sim PBD & Ours \\
\midrule
Beige hoodie     & Fold  & 0.049 & \textbf{0.045} & 0.054 \\
                 & Grasp & 0.242 & 0.231 & \textbf{0.174} \\
Blue dress       & Fold  & 0.058 & 0.040 & \textbf{0.021} \\
                 & Grasp & 0.165 & 0.129 & \textbf{0.070} \\
Brown coat       & Fold  & 0.070 & 0.061 & \textbf{0.029} \\
                 & Grasp & 0.155 & 0.102 & \textbf{0.064} \\
Green T-shirt    & Fold  & 0.046 & 0.053 & \textbf{0.025} \\
                 & Grasp & 0.239 & 0.073 & \textbf{0.058} \\
Grey pleat skirt & Fold  & 0.039 & 0.027 & \textbf{0.011} \\
                 & Grasp & 0.145 & 0.100 & \textbf{0.016} \\
White cake skirt & Fold  & 0.054 & 0.034 & \textbf{0.019} \\
                 & Grasp & 0.222 & 0.149 & \textbf{0.117} \\
White shirt      & Fold  & 0.080 & 0.039 & \textbf{0.027} \\
                 & Grasp & 0.228 & 0.153 & \textbf{0.069} \\
\midrule
Mean             & Fold  & 0.056 & 0.043 & \textbf{0.027} \\
                 & Grasp & 0.199 & 0.134 & \textbf{0.081} \\
                 & All   & 0.128 & 0.088 & \textbf{0.054} \\
\bottomrule
\end{tabular}
\end{table}

Our simulator has the lowest error in 13 of the 14 cells and reduces the mean error of Isaac Sim
PBD by 39\% and of PyBullet by 58\%. The gain is largest on grasp-and-lift, where the error is
dominated by whether the garment is held: when a grasp fails or slips, the simulated garment stays
on the table while the real one is lifted. The remaining error concentrates on thick garments such as the beige hoodie, the only cell where
another simulator is better: the real jaws stop on a bunched stack of fabric, while the simulated
mesh is a single thin layer.

\end{document}